%% file: acl_latex.tex
\pdfoutput=1
\documentclass[11pt]{article}
\usepackage{float}
\usepackage{acl}
\usepackage[utf8]{inputenc}

\usepackage{pgfplots}
\pgfplotsset{width=10cm,compat=1.9}
\usepackage[]{algpseudocode}
\usepackage[]{algorithm}
\algtext*{EndFor}%
\algtext*{EndProcedure}%
\DeclareUnicodeCharacter{2212}{−}
\usepgfplotslibrary{groupplots,dateplot}
\usetikzlibrary{patterns,shapes.arrows}
\pgfplotsset{compat=newest}

\usepackage{tikzscale} 
\usepackage{xcolor}
\usepackage{amsmath}
\usepackage{colortbl}
\usepackage{xurl}
\usepackage{bbm}
\newcommand{\probP}{\text{I\kern-0.15em P}}
\usepackage{color}
\usepackage{amssymb}
\usepackage{pifont}
\usepackage{tabularx,booktabs}
\usepackage{enumitem}
\usepackage{placeins}
\usepackage[normalem]{ulem}
\usepackage{ragged2e}
\usepackage{cleveref}
\usepackage{microtype}
\usepackage{dsfont}
\usepackage{scalerel,xparse}
\usepackage{times}
\usepackage{latexsym}
\usepackage{adjustbox}
\usepackage{subfigure}
\usepackage{amsmath}
\usepackage{amssymb} 
\usepackage{bbm}
\usepackage{ragged2e}

\usepackage{CJKutf8}
\usepackage[T1]{fontenc}
\usepackage[russian,english]{babel}

\usepackage{soul}
\sethlcolor{orange!50}

\usepackage{multirow}
\usepackage{fontawesome5}
\usepackage{xspace}
\usepackage{todonotes}
\usepackage[most]{tcolorbox}
\usepackage[table]{xcolor}
\usepackage{array}
\usepackage{colortbl}
\usepackage{makecell}
\usepackage{graphicx}
\usepackage{kotex}
\usepackage{CJKutf8}
\usepackage{xcolor}

\usepackage{subcaption}
\usepackage{caption}
\usepackage{pifont}
\usepackage{amssymb}

\input{xling-write}
\usepackage{subfigure}

\definecolor{bggray}{rgb}{0.95, 0.95, 0.95}
\usepackage[%
    framemethod=tikz,
    skipbelow=\topskip,
    skipabove=\topskip
]{mdframed}
\mdfsetup{%
    leftmargin=0pt,
    rightmargin=0pt,
    backgroundcolor=bggray,
    middlelinecolor=black,
    roundcorner=3
}
\newtcolorbox[list inside=prompt,auto counter,number within=section]{prompt}[1][]{
    colbacktitle=black!60,
    fonttitle=\small,
    coltitle=white,
    fontupper=\footnotesize,
    boxsep=4pt,
    left=0pt,
    top=0pt,
    bottom=0pt,
    boxrule=1pt,
    #1,
}

\title{\raisebox{-0.2em}{\includegraphics[height=1.1em]{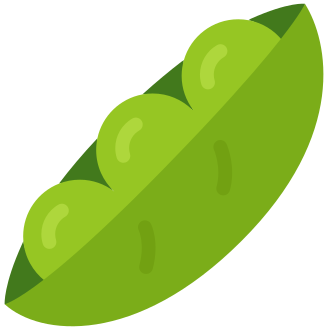}} AdaMame: A Training Recipe for Adaptive Multilingual Reasoning}
\author{\textbf{Dayeon Ki} \textsuperscript{\raisebox{-0.2em}{\includegraphics[height=1em]{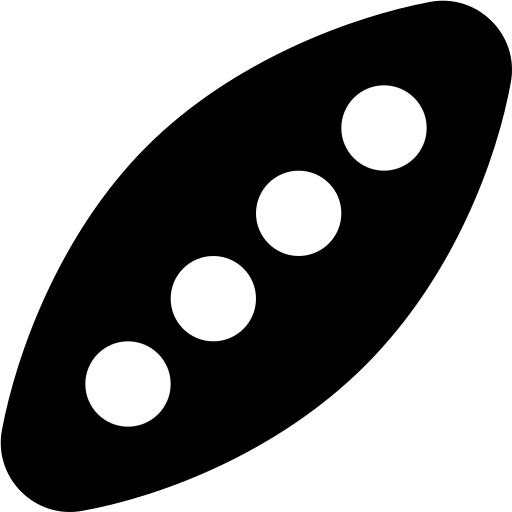}}}, 
\textbf{Kevin Duh} \textsuperscript{\raisebox{-0.2em}{\includegraphics[height=1em]{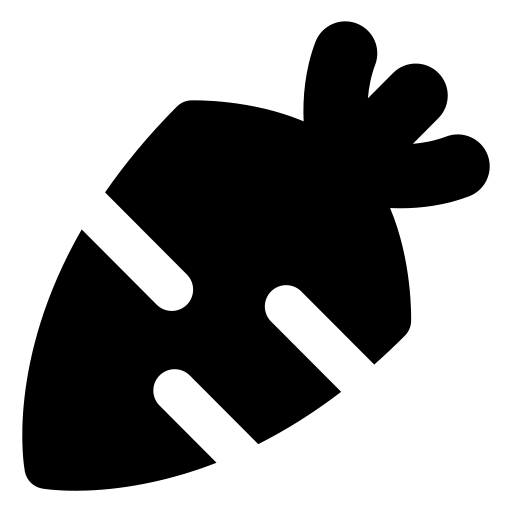}}}, 
\textbf{Marine Carpuat} \textsuperscript{\raisebox{-0.2em}{\includegraphics[height=1em]{figures/logo/peapod.png}}} \\
\textsuperscript{\raisebox{-0.2em}{\includegraphics[height=1em]{figures/logo/peapod.png}}} University of Maryland, \textsuperscript{\raisebox{-0.2em}{\includegraphics[height=1em]{figures/logo/carrot.png}}} Johns Hopkins University \\
\texttt{dayeonki@umd.edu}
}

\begin{document}
\maketitle

% Abstract
\begin{abstract}
While Large Reasoning Models (LRMs) show strong performance in English, they often fail to reason in the language of the query, a phenomenon known as language collapse. 
Existing RL-based fixes typically add a binary language fidelity reward to the accuracy objective, yet still incur trade-off in accuracy, mid-trace code-switching, and excessive token usage.
In this work, we propose \textbf{AdaMame},\footnote{Stands for \textbf{Ada}ptive \textbf{m}ultilingu\textbf{a}l \textbf{m}odel r\textbf{e}asoning.} a two-stage training recipe for multilingual mathematical reasoning that addresses these limitations by adaptively aligning the reasoning language to the query language without compromising accuracy.
The first SFT stage fine-tunes on non-MT reasoning traces across five languages to establish multilingual reasoning capability. 
In the subsequent RL stage, we introduce AdaMame-GRPO, an adaptation of Group Relative Policy Optimization (GRPO) in which a query-conditioned alignment factor grows progressively during training, guiding the model to first explore diverse reasoning languages before exploiting reasoning in the query language.
Evaluated across two benchmarks, two LRMs, and 12 languages, AdaMame-GRPO achieves Pareto-optimal performance across reasoning accuracy, language fidelity, and token efficiency over all baselines, with the strongest gains on out-of-domain, lower-resource languages.\footnote{Code, data, and model available at \url{https://github.com/dayeonki/adamame}.}
% Analysis further reveals a natural explore-then-exploit curriculum during training and shows that the alignment factor enables a controllable accuracy-fidelity trade-off.

\end{abstract}

\input{page/00_introduction}
\input{page/01_related_work}

% \input{page/01_2_preliminary}
\input{page/02_method}
\input{page/03_experimental_setup}
\input{page/04_results}
\input{page/06_conclusion}
\input{page/07_limitation}
\input{page/09_acknowledgement}

\bibliography{custom}

\clearpage
% Appendix
\appendix
\section*{Appendix}
\section{Prompts}
\label{appendix:prompt}

Figure~\ref{fig:prompt} shows the language-specific prompt instructions used for the \textbf{Prompt} and \textbf{SFT} baselines.
We follow provider-recommended prompting practices to standardize the output format \citep{deepseekai2025deepseekr1incentivizingreasoningcapability, yang2025qwen3technicalreport}.

\input{figures/prompt}

\section{Components of AdaMame}
\label{appendix:more_results}

\subsection{Non-MT Reasoning Traces}
\label{appendix:naturally}

We compare SFT on non-MT occurring reasoning traces against the conventional choice of machine-translated English traces~\citep{sutawika2026gainedtranslationprivilegedpairwise, gurgurov2026reasonxlshiftingllmreasoning}.
Specifically, we evaluate two variants: (1) non-MT reasoning traces only ($\mathcal{D}_\text{sft}$, 30K), and (2) non-MT traces augmented with 65K machine-translated traces from~\citet{sutawika2026gainedtranslationprivilegedpairwise}, which are sourced from DeepScaleR~\citep{deepscaler2025} and translated using \textsc{GPT-5 nano}.
Despite training on over three times more data (95K vs.\ 30K), augmenting with machine-translated traces degrades both reasoning accuracy and LCPR, confirming our choice to use non-MT traces exclusively in all main experiments.

\input{tables/naturally}

\subsection{LoRA vs. Full SFT}
\label{appendix:lora}

We compare SFT with LoRA and full fine-tuning for the \textsc{Distill-Qwen 1.5b} backbone, with LoRA applied to all linear modules.
Training corpora $\mathcal{D}_\mathrm{sft}$ and all hyperparameters are held constant, except the learning rate, which is reduced from $2\mathrm{e}^{-4}$ to $2\mathrm{e}^{-5}$ for full fine-tuning \citep{bohnet2025comparative}. 
As shown in Table~\ref{tab:lora}, LoRA yields higher reasoning accuracy and LCPR on MGSM-Rev2, and we therefore adopt it in all SFT experiments.

\input{tables/lora}

\subsection{Accuracy vs. Accuracy+Format Reward}
\label{appendix:format}

While standard GRPO uses a binary accuracy reward alone, several prior works augment it with a format reward that penalizes malformed rollouts~\citep{deepseekai2025deepseekr1incentivizingreasoningcapability, rastogi2025magistral}.
We define the format reward as:
\begin{equation}
r_\mathrm{format} = 
\begin{cases}
1, & \text{if format is correct} \\
0, & \text{if format is incorrect}
\end{cases}
\end{equation}
and compare it against the accuracy-only reward in Table~\ref{tab:format}. We show that accuracy-only reward yields higher reasoning accuracy and LCPR than the combined reward, and we therefore adopt it in all main experiments.

\input{tables/format}

\subsection{Scheduler Variant}
\label{appendix:scheduler}

We compare cosine growth scheduler against two alternatives: (1) threshold-based scheduler, which increases the query alignment weight whenever language fidelity falls below a pre-defined threshold (80.0\%) among the rollouts, and (2) Lagrangian multiplier scheduler, which treats language fidelity as a constraint, maximizing $\mathrm{E}$[reasoning accuracy] subject to $\mathrm{E}$[language fidelity] greater than or equal to 80.0. As shown in Table~\ref{tab:scheduler}, our cosine growth scheduler outperforms both alternatives in terms of reasoning accuracy and LCPR. We therefore use this scheduler in our main experiments.

\input{tables/scheduler}

\subsection{GRPO vs. Dr.GRPO}
\label{appendix:dr_grpo}

Dr.GRPO is a variant of GRPO that removes the length-normalization and standard deviation terms from the advantage computation, mitigating pre-training and optimization biases that otherwise inflate response length during training \citep{liu2025understandingr1zeroliketrainingcritical}.
Given its adoption in recent multilingual reasoning work~\citep{sutawika2026gainedtranslationprivilegedpairwise, gurgurov2026reasonxlshiftingllmreasoning}, we evaluate both GRPO and AdaMame-GRPO on top of both standard GRPO and Dr.GRPO (Table~\ref{tab:dr_grpo}). For both GRPO and AdaMame-GRPO, using Dr.GRPO yields higher reasoning accuracy with comparable LCPR, and we therefore adopt it in all main RL experiments.

\input{tables/dr_grpo}

\subsection{Sampling Strategies}
\label{appendix:sampling}

We compare three data sampling strategies for constructing the 5K corpus $\mathcal{D}_\mathrm{grpo}$: 
\begin{itemize}[leftmargin=*, itemsep=2pt, parsep=-1pt]
    \item \textbf{Random sampling}: queries are randomly sampled from $\mathcal{D}_\mathrm{sft}$ without filtering.
    
    \item \textbf{Conditional sampling}: one candidate is sampled per query; a query is retained if the rollout is in the query language but leads to an incorrect answer (accuracy as 0, language fidelity as 1.0).
    
    \item \textbf{Rejection sampling}: 8 candidates are generated per query; a query is retained if the model produces both correct and incorrect rollouts (i.e., $0 < |\mathcal{O}_\mathrm{correct}| < 8$).
\end{itemize}
As shown in Table~\ref{tab:sampling}, rejection sampling yields the highest reasoning accuracy and LCPR, and we therefore adopt it in all main experiments.

\input{tables/sampling}

\subsection{Distribution of $|\mathcal{O}_\mathrm{correct}|$}
\label{appendix:distribution}

In Table~\ref{tab:distribution}, we report the distribution  of $|\mathcal{O}_\mathrm{correct}|$. The weighted mean of $|\mathcal{O}_\mathrm{correct}|$ is 2.75 across the 5K RL dataset. This indicates that the queries used in our RL step are, on average, relatively difficult.

\input{tables/distribution}

\subsection{Language Detection Reliability}
\label{appendix:reliability}

As noted in Limitations section, AdaMame-GRPO depends on the reliability of the language detector used to compute the query alignment reward.
We validate the \textsc{lingua} detector on two settings that mirror its roles in our pipeline: (1) short-context detection on MGSM-Rev2 queries with gold language labels, corresponding to $\mathrm{lang}(q)$, and (2) long-context detection on randomly sampled 200 reasoning traces from mCoT-MATH \citep{lai-nissim-2024-mcot}, corresponding to $\mathrm{lang}(o_i)$.
As shown in Table~\ref{tab:lang_detection}, \textsc{lingua} detector achieves 100\% accuracy on short-context detection and 99.2\% on long-context detection, demonstrating strong reliability across both settings.

\input{tables/lang_detection}

\section{Experiment Setup Details}

\subsection{Dataset Construction}
\label{appendix:dataset}

\input{tables/filtering}

We prepend the language-specific instructions from Appendix~\ref{appendix:prompt} when generating reasoning traces with \textsc{GPT-5 nano}, using a sampling temperature of 1.0 for diversity.
Table~\ref{tab:filtering} reports per-language retain rates after filtering, with an average of 72.2\%, with Thai having the highest rate (73.5\%) and Korean the lowest (69.2\%).

\subsection{Reasoning Trace Statistics}
\label{appendix:statistics}

We three additional analyses on the generated multilingual reasoning traces: (1) the length of the generated reasoning traces, measured in number of characters; and, for a random sample of 1K traces per in-domain language (5K in total), LLM-as-a-judge evaluation using \textsc{GPT-5.5} for (2) faithfulness (i.e., is the reasoning trace faithful to the final answer?) and (3) error rate (i.e., does the reasoning trace contain any hallucinated or erroneous intermediate steps?). For (2) and (3), we obtain a binary label (Yes/No) and compute the proportion of ``Yes'' responses. We show results in Tables~\ref{tab:stat1} and \ref{tab:stat2}.

\input{tables/stat1}
\input{tables/stat2}

% \subsection{Training Strategies}
% \label{appendix:train_details}

% \paragraph{SFT.}
% We adopt LoRA \cite{hu2021loralowrankadaptationlarge} for parameter efficient fine-tuning targeting all linear modules, implemented using \textsc{LLaMA-Factory} \citep{zheng-etal-2024-llamafactory}. 
% We set LoRA rank $r$ as 8, $\alpha$ as 16, cutoff length as 8,192 tokens, learning rate as $2\mathrm{e}^{-4}$ with cosine scheduler and warm-up ratio 0.1.
% We use \texttt{deepseekr1} template for \textsc{Distill-Qwen 1.5b} and \texttt{qwen3\_think} for \textsc{Qwen-3 4b}
% % , and \texttt{smollm3\_think} for \textsc{SmolLM3 3b}.
% One run of LoRA SFT requires around 1.5 hours on one NVIDIA-A5000 GPU.

% \paragraph{RL.}
% We implement all RL experiments in \textsc{verl} \citep{verl}, with modifications to support the AdaMame-GRPO variant.
% We set max batch size to 512, mini batch size to 256 (2 gradient updates per batch), rollout count to 8, rollout temperature to 0.8, learning rate $1\mathrm{e}^{-5}$, max prompt length to 512, max response length to 4,096, KL loss coefficient to 0.001, and the number of training epochs is set to 10.
% We find the optimal configuration for the learning rate, temperature, and KL loss coefficient using \textsc{wandb}'s sweep search.\footnote{\url{https://docs.wandb.ai/models/ref/python/functions/sweep}}
% A single GRPO run takes approximately 4 hours on one NVIDIA-A100 GPU for \textsc{Distill-Qwen 1.5b}, and approximately 10 hours on two NVIDIA-A100 GPUs for \textsc{Qwen3 4b}. 

\subsection{MGSM-Rev2 Details}
\label{appendix:mgsm_query}

In Table~\ref{tab:mgsm_details}, we report the proportion of MGSM queries revised in MGSM-Rev2 during the process of translation quality and ambiguity correction.

\input{tables/mgsm_details}

\subsection{Evaluation Metric Details}
\label{appendix:metric_details}

We provide implementation details of the LCPR (Language Confusion Pass Rate) metric. We follow the procedure introduced in \citet{marchisio-etal-2024-understanding} and use the \textsc{lingua} detector for detecting language(s) of a text. Given a reasoning trace $c$,

\paragraph{Line-level Pass Rate (LPR).}
We split $c$ into lines (by newline character) and check against the query language $q_\ell$. LPR is the percentage of reasoning traces where all lines match $q_\ell$:
\begin{equation}
\mathrm{LPR} = \frac{|\mathcal{C} \textbackslash \mathcal{C}_{\neg \ell}|}{|\mathcal{C}|},
\end{equation}
where $\mathcal{C}$ is the set of all reasoning traces and $\mathcal{C}_{\neg \ell}$ is the set of traces that contain line-level errors.

\paragraph{Word-level Pass Rate (WPR).}
We first exclude traces with line-level errors ($\mathcal{C}_{\neg \ell}$, as most line-level errors would also be counted toward word-level errors, making it difficult to distinguish between the two error types \citep{marchisio-etal-2024-understanding}.
For languages that use Latin script, we identify characters outside of the script's Unicode range.
For languages that do not use Latin script, we detect errorneous English words (since languages mostly code-switch with English) that do not typically occur in target language text.
WPR is the percentage of reasoning traces where all words are in $q_\ell$:
\begin{equation}
\mathrm{WPR} = \frac{|(\mathcal{C} \textbackslash \mathcal{C}_{\neg \ell}) \textbackslash \mathcal{C}_{\neg w}|}{|\mathcal{C} \textbackslash \mathcal{C}_{\neg \ell}|},
\end{equation}
where $\mathcal{C}_{\neg w}$ is the set of traces that contain word-level errors.

\paragraph{Language Confusion Pass Rate (LCPR).}
LCPR is a harmonic mean of line-level and word-level pass rates:
\begin{equation}
\mathrm{LCPR} = 2 \times \frac{\mathrm{LPR} \times \mathrm{WPR}}{\mathrm{LPR} + \mathrm{WPR}}
\end{equation}

\subsection{Language Consistency vs. LCPR}
\label{appendix:cs}

The language consistency (i.e., fidelity) metric used in prior work~\citep{zhang2026thinknativelyunlockingmultilingual, gao2026explangimprovedexplorationexploitation} measures fidelity by the top-1 detected language alone, treating a trace as fully language-faithful as long as its dominant language matches the query.
LCPR, by contrast, computes the harmonic mean of line- and word-level language pass rates (LPR and WPR), capturing code-switching that top-1 detection overlooks.
As shown in Table~\ref{tab:lang_consistency}, a reasoning trace can receive a perfect language consistency score of 1.0 while exhibiting substantial code-switching, which LCPR correctly penalizes.

\input{tables/models}
\input{tables/datasets}
% \input{tables/comet_qe}
\input{tables/languages}
\input{tables/lang_consistency}

\section{Detailed Results}
\label{appendix:results}

\subsection{Code-Switching Rate}
\label{appendix:code-switch}

We adopt a straightforward operational definition of ``code-switching'' as the presence of more than one language within a reasoning trace. We compute a code-switching rate as the fraction of traces in which this is detected using the \textsc{lingua} language detector. 
As shown in Table~\ref{tab:code-switch}, both \textit{M-Thinker} baselines exhibit the highest code-switching rates among all variants, with more than twice the rate observed for AdaMame-GRPO.

\input{tables/code-switch}

\subsection{Per-Language Results}
\label{appendix:per_lang}

We report per-language results for all models and evaluation datasets in Table~\ref{tab:per_lang_1} and Table~\ref{tab:per_lang_2}.

\subsection{Length Distribution Comparison}
\label{appendix:length}

We compare the length distribution of the SFT training dataset generated by \textsc{GPT-5 nano} against the reasoning traces produced by the baselines in Table~\ref{tab:main_results}, in terms of number of characters. 
As shown in Table~\ref{tab:length}, the average length of the training dataset is 903.4, compared to 297.4 for AdaMame-GRPO. This shows that while the \textsc{GPT-5 nano}-generated training traces are already shorter than those from the Vanilla baseline, AdaMame-GRPO-generated traces are the shortest on average, highlighting the token efficiency of the proposed method. 

\input{tables/length}

\subsection{Change in $\beta$ Results}
\label{appendix:change_in_b}

We report numerical results for varying the query alignment factor in AdaMame-GRPO in Table~\ref{tab:change_in_b}.

\input{tables/change_in_b}

\input{tables/per_lang_1}
\input{tables/per_lang_2}

\section{Usage of Large Language Models}
We used LLMs to support and refine the writing of our work, such as for stylistic adjustments, including improving readability and removing layout issues (e.g., widows and orphans).

\end{document}

%% file: xling-write.tex
\definecolor{zoey green}{rgb}{0.684,0.836,0.227}
\newcommand{\ignore}[1]{}

%% file: page/00_introduction.tex
\section{Introduction}

% Multilingual reasoning suggested to improve reasoning accuracy and language fidelity 
    % Why language-specific reasoning is important?
    % language collapse issue 
% Large Reasoning Models (LRMs) deployed in multilingual settings must satisfy two objectives simultaneously: producing correct answers (reasoning accuracy) and generating reasoning traces (i.e., chains of thoughts, CoTs) in the same language as the query (language fidelity) \citep{shi2022language, muennighoff2023crosslingual, yong2025crosslingual}.
% Language fidelity matters for both practical and technical reasons. 
% Practically, users who interact with models in their native language expect responses in kind, and the absence of these can hurt their understanding of and trust in models \citep{aggarwal2025languagemodelsfactualitydepends}.
% Technically, effective reasoning strategies can vary across languages \citep{ki2026makesgoodmultilingualreasoning, gurgurov2026reasonxlshiftingllmreasoning}, models can sometimes reason more effectively in the original query language \citep{gao2025thinkingmultilinguallyempowerllm}, and multilingual thinking promotes output diversity \citep{blasi2022over, xu2026languagethoughtshapesoutput}. 
% Yet because the vast majority of LRM training data is in English \citep{ghosh-etal-2025-survey}, these models suffer from the so-called language collapse issue, where models default to reasoning in English regardless of query language \citep{park2026crosslingualcollapselanguagecentricfoundation}.

Large Reasoning Models (LRMs) deployed in multilingual settings are typically evaluated along two dimensions: producing correct answers (reasoning accuracy) and generating reasoning traces (i.e., chains of thoughts, CoTs) in the same language as the user query (language fidelity) \citep{shi2022language, muennighoff2023crosslingual, yong2025crosslingual}.

Language fidelity matters in settings where reasoning traces are visible to end users in support of downstream decision-making. 
Practically, it can support consistency between the user's input and the model's observable reasoning process thus give users greater transparency into model behavior. For users interacting in their native language, reasoning traces in an unfamiliar language may also reduce their ability to understand and trust the model's responses \citep{aggarwal2025languagemodelsfactualitydepends}.
Technically, effective reasoning strategies can vary across languages \citep{ki2026makesgoodmultilingualreasoning}, models can sometimes reason more effectively in the user query language \citep{gao2025thinkingmultilinguallyempowerllm}, and multilingual thinking promotes output diversity \citep{blasi2022over, xu2026languagethoughtshapesoutput}. Yet because the vast majority of LRM training data is in English \citep{ghosh-etal-2025-survey}, these models suffer from the so-called language collapse issue, where models default to reasoning in English regardless of query language \citep{park2026crosslingualcollapselanguagecentricfoundation}.

\input{figures/teaser_fig_vis}

\input{figures/main_fig_vis}

% While some efforts in improving this, they are limited
    % Challenges of prior work => as shown in Figure 1a
Recent efforts to remedy language collapse share a common design of appending a binary language fidelity reward to the accuracy objective through a manually tuned weighting ratio \citep{zhang2026thinknativelyunlockingmultilingual, sutawika2026gainedtranslationprivilegedpairwise, gao2026explangimprovedexplorationexploitation}.
Despite its appeal, this approach has several persistent limitations (Figure~\ref{fig:main_fig}): (1) a trade-off in reasoning accuracy \citep{wang2025polymath}; (2) alternating languages (i.e., code-switching) within reasoning traces \citep{wang-etal-2025-language-mixing}; and (3) overthinking, where models spend excessive tokens on reasoning without proportional gains \citep{chen2024not, sui2025stop}.
Most existing methods also require English reference reasoning traces \citep{sutawika2026gainedtranslationprivilegedpairwise, zhang2026thinknativelyunlockingmultilingual} and depend on a fixed, developer-specified weighting regime, which limits scalability \citep{gurgurov2026reasonxlshiftingllmreasoning}.

% In this work, we propose AdaMame, training recipe capable of adaptively selecting reasoning language based on query language, balancing both reasoning accuracy and language fidelity. 
% Training => two-stage training framework, stage 1: SFT to equip fundamental cpaability of reasoning in multiple language, stage 2: introduce AdaMaMe, adaptation of GRPO inspired by ARM, which addresses two issues: (1) increased reasoning accuracy, (2) matching to query language
    % Adamame => query alignment weight, which grow
In this work, we propose \raisebox{-0.2em}{\includegraphics[height=1.1em]{figures/logo/bean.png}} \textbf{AdaMame}, a two-stage training recipe for multilingual mathematical reasoning that adaptively aligns the reasoning language to the query language, jointly optimizing for reasoning accuracy and language fidelity (\S\ref{sec:method}).
AdaMame builds on the well-established SFT-then-RL post-training recipe, introducing targeted modifications to shift model behavior toward the query language.
In Stage 1, we apply supervised fine-tuning (SFT) on non-machine-translation (MT) reasoning traces in five languages, equipping the model with foundational multilingual reasoning capability and sensitivity to language-specific reasoning patterns \citep{ki2026makesgoodmultilingualreasoning}.
In Stage 2, we introduce \textbf{AdaMame-GRPO}, an adaptation of Group Relative Policy Optimization (GRPO) \citep{shao2024deepseekmath} inspired by ARM \citep{wu2026arm}, that incorporates a query-conditioned alignment factor which adaptively grows during training. This encourages the model to progressively align its reasoning language to the query while preserving reasoning accuracy as the main objective.

% Our results show that training 3 LRMs with AdaMame achieve better performance, with lower language confusion rate (as shown in Figure 1b), and with  better token efficiency than baselines, across both in-domain and out-of-domain languages in matehmatical reasoning tasks.
Across two multilingual mathematical reasoning benchmarks, two LRMs, and 12 in-domain and out-of-domain languages (\S\ref{sec:setup}), we show that AdaMame-GRPO achieves Pareto-optimal performance across accuracy and language fidelity while using fewer tokens than all baselines.
While the SFT stage alone substantially reduces language collapse, the RL stage with AdaMame-GRPO further improves generalization to out-of-domain languages, with the largest gains on low-resource languages (\S\ref{sec:results}). 
Further analysis confirms that AdaMame-GRPO induces the intended explore-then-exploit curriculum: models initially explores diverse reasoning languages before progressively converging on the query language as the query alignment factor grows (\S\ref{sec:analysis_1}), and increasing the weight of the alignment factor yields improved language fidelity with a controllable accuracy trade-off (\S\ref{sec:analysis_2}).

% Our additional analysis shows that (1) AdaMame effectively gradually aligns to the query langauge throughout the GRPO process, (2) leading to better language fidelity to the query language at the end of training, (3) increasing the weight of the query alignment component has diminishing returns, trade-off between accuracy and LCPR

% In summary, our contributions are three-fold: (1) We introduce ADamame, a training receipe for multilingual reasoning that improves both reasoning accuracy and language fidelity, (2) We show that this variant outperforms baselines, with better token efficiency, larger improvements on lower-resource languages + better generalization to unseen languages, (3) Increasing B gives with a trade-off, language alignment ratio grows over training process ->exploriation
In summary, our contributions are three-fold:
\begin{itemize}[leftmargin=*, itemsep=2pt, parsep=-1pt]
    \item We introduce AdaMame, a two-stage post-train- ing recipe (SFT+RL) for multilingual mathematical reasoning, with targeted modifications for query language alignment.
    
    \item We release a dataset of 30K non-MT reasoning traces across five languages (French, Portuguese, Japanese, Korean, and Thai), providing a valuable resource for post-training multilingual reasoning models \citep{barua2026longchainofthoughtreasoninglanguages}.
    
    \item AdaMame-GRPO achieves Pareto-optimal performance across reasoning accuracy, language fidelity, and token efficiency over all tested baselines, with the strongest generalization to out-of-domain and lower-resource languages.
\end{itemize}

%% file: figures/teaser_fig_vis.tex
\begin{figure}
    \centering
    % \begin{minipage}[t]{0.75\linewidth}
    %     \centering
    %     \includegraphics[width=\linewidth]{figures/main_fig.pdf}
    %     \caption*{(a) Comparison of reasoning behaviors for General Model vs.\ AdaMame.}
    % \end{minipage}
    % \hfill
    % \begin{minipage}[t]{0.24\linewidth}
    %     \centering
    %     \includegraphics[width=\linewidth]{figures/teaser_fig.pdf}
    %     \caption*{(b) Accuracy vs.\ LCPR}
    % \end{minipage}
    \includegraphics[width=0.9\linewidth]{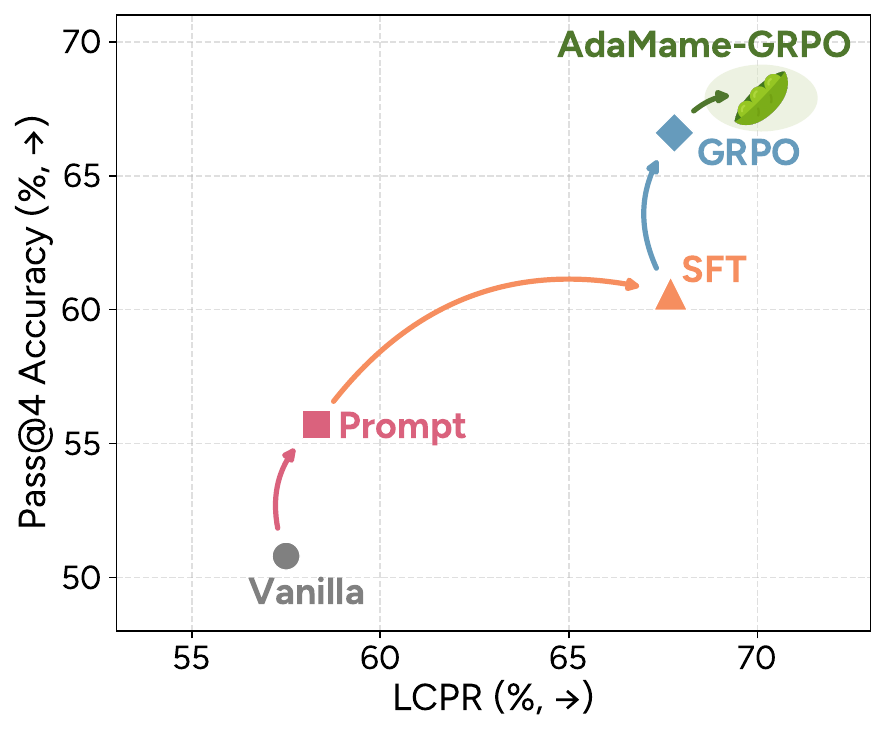}
    \caption{\textbf{Accuracy (Pass@4) versus Language Fidelity (LCPR: Language Confusion Pass Rate).}
    Backbone model: \textsc{Distill-Qwen 1.5b}.
    AdaMame-GRPO achieves a Pareto-optimal performance on both objectives.}
    \label{fig:teaser_fig}
\end{figure}

    % (\textbf{b}) Pass@4 accuracy versus LCPR (Language Confusion Pass Rate) across model variants of \textsc{Distill-Qwen 1.5b}. AdaMame occupies the upper-right, reflecting its effectiveness in both reasoning accuracy and language fidelity.}

%% file: figures/main_fig_vis.tex
\begin{figure*}
    \centering
    % \begin{minipage}[t]{0.75\linewidth}
    %     \centering
    %     \includegraphics[width=\linewidth]{figures/main_fig.pdf}
    %     \caption*{(a) Comparison of reasoning behaviors for General Model vs.\ AdaMame.}
    % \end{minipage}
    % \hfill
    % \begin{minipage}[t]{0.24\linewidth}
    %     \centering
    %     \includegraphics[width=\linewidth]{figures/teaser_fig.pdf}
    %     \caption*{(b) Accuracy vs.\ LCPR}
    % \end{minipage}
    \includegraphics[width=\linewidth]{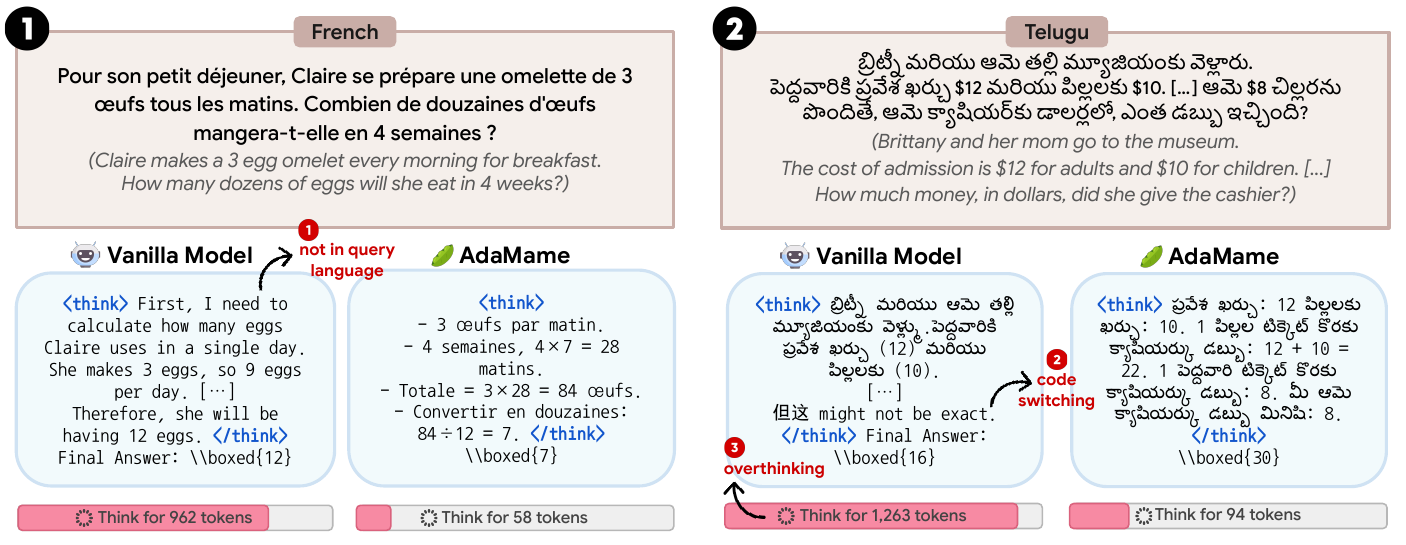}
    \caption{\textbf{Effectiveness of AdaMame.} We compare reasoning behaviors between a Vanilla Model and AdaMame for French and Telugu queries. While the Vanilla Model fails to answer correctly in the query language, code-switches mid-trace, and overthinks with excessive token usage, AdaMame adapts to the query language, producing both correct and token-efficient reasoning.}
    \label{fig:main_fig}
\end{figure*}

    % (\textbf{b}) Pass@4 accuracy versus LCPR (Language Confusion Pass Rate) across model variants of \textsc{Distill-Qwen 1.5b}. AdaMame occupies the upper-right, reflecting its effectiveness in both reasoning accuracy and language fidelity.}

%% file: page/01_related_work.tex
\input{tables/comparison}

\section{Related Work}

\subsection{Improving Multilingual Reasoning}

A growing body of work documents substantial performance gaps \citep{luo-etal-2025-mmath, zhao-etal-2026-comprehensive} and language collapse in LRMs when queries are posed in languages other than English \citep{park2026crosslingualcollapselanguagecentricfoundation}.
Prior methods to address these issues fall into two directions. 
First is prompt-based methods, such as adding language-specific instructions \citep{yong2025crosslingual, qi-etal-2025-models} or prefixes at the beginning of model generations \citep{tam2025languagemattersmultilingualinput} to steer the output language.
Second is fine-tuning models on multilingual CoTs \citep{lai-nissim-2024-mcot, she2024mapo, chai2025xcot}.

More recent Reinforcement Learning (RL)-based approaches augment the GRPO reward with an explicit language fidelity term \citep{liu2026selfimprovingmultilinguallongreasoning, zhang2026thinknativelyunlockingmultilingual, sutawika2026gainedtranslationprivilegedpairwise}. 
For instance, M-Thinker combines two reward terms in addition to the standard GRPO: (1) a binary language fidelity reward that fires when the reasoning language matches the query, and (2) a cross-lingual thinking alignment reward where an LLM-as-judge scores how closely the reasoning trace aligns with a reference English trace \citep{zhang2026thinknativelyunlockingmultilingual}.

While effective, these methods exhibit three persistent failure modes (Figure~\ref{fig:main_fig}): (1) trading off answer correctness for matching reasoning language to the query language, (2) code-switching within reasoning traces, and (3) overthinking with excessive use of tokens.
As compared in Table~\ref{tab:comparison}, many prior approaches also require English reference reasoning traces for supervision and rely on a fixed, developer-specified weighting regime between reward components \citep{gao2026explangimprovedexplorationexploitation, gurgurov2026reasonxlshiftingllmreasoning}, limiting robustness to changes in that ratio.
AdaMame-GRPO addresses this gap with an \textit{adaptive} reward that progressively strengthens query language alignment over the course of GRPO training, requiring no English reference traces.

\subsection{Adaptive Reasoning}
Recent studies have explored making LRM reasoning adaptive, ranging from binary decisions about whether to engage in thinking \citep{tu2026learning} to fine-grained adaptation of reasoning format or effort based on task difficulty \citep{yu2025think, wu2026arm, wang2026adareasoner, yang2026aresadaptivereasoningeffort}.

In multilingual settings, adaptive reasoning has been studied to select whichever reasoning language is most effective for a given query, through a language router \citep{guo2026learning}, LLM-as-judge \citep{zheng2025adamcotrethinkingcrosslingualfactual}, or comparison against English traces \citep{ye2026x1learningthinkadaptively}.
AdaMame-GRPO is distinctive among these approaches in that it does not treat reasoning language as a free variable to be selected per query.
Instead, it fixes query language alignment as the objective, and progressively adapt to this objective throughout GRPO training by increasing the query alignment factor.
% AdaMame shares the goal of adaptive reasoning language selection, but is distinctive in targeting alignment with the query language rather than unconstrained accuracy-maximizing language choice. 

%% file: tables/comparison.tex
\begin{table*}
\centering
\resizebox{\linewidth}{!}{%
    \begin{tabular}{l l l}
    \toprule
    \textbf{Method} & \textbf{Dataset} & \textbf{Objective} \\
    \toprule

    \textbf{M-Thinker} \citep{zhang2026thinknativelyunlockingmultilingual} & $q, c, o, c_{\mathrm{en}}$ & GRPO (Accuracy + Language Fidelity + Format + Cross-lingual Thinking Alignment) \\
    \textbf{SP3F} \citep{sutawika2026gainedtranslationprivilegedpairwise} & $q, c, o, c_{\mathrm{en}}$ & GPRO (Accuracy + Language Fidelity + Format + Judge Preference Feedback) \\
    \textbf{TRIT} \citep{liu2026selfimprovingmultilinguallongreasoning} & $q,c,o$ & GRPO (Accuracy + Language Fidelity + Format + Repetition Penalty) \\
    \textbf{ExpLang} \citep{gao2026explangimprovedexplorationexploitation} & $q, c, o$ & GRPO (Accuracy + Pass@k + Language Fidelity + Format + Thinking Language Diversity) \\
    \textbf{ReasonXL} \citep{gurgurov2026reasonxlshiftingllmreasoning} & $q,c,o$ & GRPO (Accuracy + Language Fidelity + Format + Repetition Penalty) \\
    \raisebox{-0.2em}{\includegraphics[height=1.1em]{figures/logo/bean.png}} \textbf{AdaMame} (\textit{Ours}) & $q,c,o$ & AdaMame-GRPO (\S\ref{sec:method_2}) \\

    \bottomrule
    \end{tabular}
}
\caption{\textbf{Comparison of AdaMame to prior approaches.} \textbf{Dataset}: required training data components; \textbf{Objective}: reward components used during RL training. 
$q$: query, $c$: reasoning trace, $o$: final output, $c_\mathrm{en}$: English reference trace. AdaMame requires no English reference traces and uses no manually tuned weighting between reward components.}
\label{tab:comparison}
\end{table*}

%% file: page/02_method.tex
\section{\raisebox{-0.2em}{\includegraphics[height=1.1em]{figures/logo/bean.png}} AdaMame: A Training Recipe}
\label{sec:method}

AdaMame builds on the SFT-then-RL post-training recipe with targeted modifications that optimizes reasoning accuracy, language fidelity, and token efficiency by adaptively aligning reasoning to the query language.
We first construct a high-quality multilingual training dataset (\S\ref{sec:method_0}), then train in two stages: SFT (\S\ref{sec:method_1}) followed by RL (\S\ref{sec:method_2}).

\subsection{Preparing Ingredients}
\label{sec:method_0}

We consider multilingual mathematical reasoning, where a model receives a query $q$ posed in some language $\ell$ and must produce a reasoning trace $c$ and a final output $o$. 
The goal is to produce correct $o$ matching the ground truth answer $g$ and $c$ in $\ell$.

Our training dataset therefore requires two components: (1) queries $q$ across all $\ell \in \mathcal{L}$, and (2) non-MT reasoning traces $c$ in the respective query language $\ell$, not the machine-translated counterparts of English traces. 
This enables models to reason across languages $\mathcal{L}$ and learn language-specific reasoning patterns (see Appendix~\ref{appendix:naturally} for details).

\paragraph{Queries.}
We sample queries from DAPO-MATH-17K \citep{yu2026dapo} across five in-domain languages (French, Portuguese, Japanese, Korean, and Thai), sourced from \citet{liu2026selfimprovingmultilinguallongreasoning}.\footnote{We use DAPO-MATH-17K since it provides various difficulty and manually verified queries \citep{yu2026dapo}.} Each $q$ is machine-translated from English using \textsc{DeepSeek-V3.2-Exp}~\citep{deepseekai2024deepseekv32}, with translation quality verified by \textsc{Qwen3 32B}~\citep{yang2025qwen3technicalreport}.

\paragraph{Reasoning Traces.}
For each query $q$, we generate reasoning trace $c$ using \textsc{GPT-5 nano} \citep{singh2026openaigpt5card}.\footnote{\textsc{GPT-5 nano} is the most cost-effective and time-efficient option among other multilingual LRMs. When compared with \textsc{Qwen-3} and \textsc{Distill-DeepSeek-Qwen} model families, we find they frequently generate English or Chinese reasoning traces despite language-specific instructions, resulting in lower retain rates than \textsc{GPT-5 nano} (72.2\% average retain rate).} We retain a triplet if: $c$ is in the same language as $q$, $c$ leads to a correct answer in $o$, and follows the required format (\texttt{<think></think>} 
for $c$, \texttt{\textbackslash\textbackslash boxed\{\}} in $o$), leading to an average retain rate of 72.2\%. Appendix \ref{appendix:dataset} provides further details on the generation and filtering process.

\subsection{Stage 1: SFT for Learning}
\label{sec:method_1}

\paragraph{Dataset.}
We leverage SFT as a cold start to expose the model to diverse reasoning languages and language-specific reasoning patterns \citep{gao2026explangimprovedexplorationexploitation, wu2026arm}.
For each of the five languages, we sample 6K triplets $(q, c, o)$ from \S\ref{sec:method_0}, yielding a 30K SFT training corpus $\mathcal{D}_\mathrm{sft}$. We report detailed statistics in Appendix~\ref{appendix:statistics}.

\paragraph{Objective.}
We take an open-weight LRM $\mathcal{M}$ and fine-tune on $\mathcal{D}_\mathrm{sft}$, resulting in $\mathcal{M}_\mathrm{sft}$.
During fine-tuning, each query $q$ is prepended with a language-specific prompt instruction (Appendix~\ref{appendix:prompt}) that explicitly encourages reasoning in the query language.
To prevent $\mathcal{M}_\mathrm{sft}$ from catastrophic forgetting and to improve training efficiency, we fine-tune with Low-rank adaptation (LoRA) \citep{hu2021loralowrankadaptationlarge}, which constrains the magnitude of parameter updates \citep{gao2026explangimprovedexplorationexploitation}. 
We show that LoRA outperforms full fine-tuning on both reasoning accuracy and language fidelity (details in Appendix~\ref{appendix:lora}).

\subsection{Stage 2: RL for Generalizing}
\label{sec:method_2}

\paragraph{Dataset.}
To construct the RL training corpus, we apply rejection sampling over $\mathcal{D}_\mathrm{sft}$  with 8 candidates per query \citep{zhang2026thinknativelyunlockingmultilingual}.\footnote{We compare three different data sampling strategies for RL corpus construction in Appendix~\ref{appendix:sampling}.}
A query $q$ is selected if the model generates both correct and incorrect rollouts (i.e., $0<|\mathcal{O}_\mathrm{correct}|<8$), selecting problems that are challenging yet solvable.\footnote{Distribution of $|\mathcal{O}_\mathrm{correct}|$ is in Appendix~\ref{appendix:distribution}.}
We then randomly sample 1K instances for each of five in-domain languages, yielding a 5K corpus $\mathcal{D}_\mathrm{grpo}$, with a held-out validation set of 1K instances.
% (250 instances per language).

% While $\mathcal{M}_\text{sft}$ learns to reason across multiple languages, it lacks the ability to adaptively align with the query language, particularly for out-of-domain languages (\S\ref{sec:results}).

\paragraph{GRPO.}
While $\mathcal{M}_\text{sft}$ learns to reason across the trained languages, it lacks the ability to generalize to out-of-domain settings compared to RL-trained models (\S\ref{sec:results}; \citet{chu2025sft}).
We therefore further train $\mathcal{M}_\text{sft}$ with standard GRPO \citep{shao2024deepseekmath}.
Here, given a query $q$ with a ground truth answer $g$, the model samples a group of $G$ outputs (i.e., rollouts) $\mathcal{O}=\{o_1, o_2, ...,o_G\}$. For each $o_i$, a binary reward $r_i$ is computed using a rule-based function that checks whether $o_i$ matches the ground truth answer $g$:\footnote{We find that accuracy-only reward outperforms a combined accuracy and format reward; see Appendix~\ref{appendix:format}.}
\begin{equation}
    r(q,o_i,g) = \mathbbm1(o_i=g).
\end{equation}
However, since this reward optimizes solely for correctness, the model defaults toward whichever language that maximizes accuracy, typically English, without any incentive to explore alternative reasoning languages or align with the query language.
Standard GRPO thus offers limited gains for language collapse (\S\ref{sec:results}).
To address this, we propose an adaptation of GRPO, which enables the model to progressively align to the query language through a query-conditioned alignment reward mechanism.

\paragraph{AdaMame-GRPO.}
Inspired by ARM \citep{wu2026arm}, we adapt GRPO by introducing a query alignment scaling factor $\alpha_i(t)$ that amplifies the base accuracy reward $r(q,o_i,g)$. Formally:
\begin{equation}
    r'(q, o_i, g) = \colorbox{yellow!30}{$\alpha_i(t)$} \cdot r(q, o_i, g),
\end{equation}
\begin{equation}
    \alpha_i(t) = 1 + \beta \cdot \mathbbm{1}\!\left[\mathrm{lang}(o_i) = \mathrm{lang}(q)\right] \cdot \phi(t),
\end{equation}
\begin{equation}
    \phi(t) = \max\!\left(\frac{1 - \cos\!\left(\pi \cdot \dfrac{t}{T}\right)}{2},\ 0.1\right),
\end{equation}
where $\mathrm{lang}(\cdot)$ denotes the detected language of a rollout or query, measured using \textsc{lingua} language detector,\footnote{\href{https://pypi.org/project/lingua-language-detector}{\texttt{lingua-language-detector}}} $\phi(t)$ is a cosine growth schedule that gradually increases from 0.1 at the beginning of training ($t=0$) to 1.0 at the end ($t=T$),\footnote{We compare cosine growth scheduler to threshold-based and Lagrangian multiplier scheduler in Appendix~\ref{appendix:scheduler}.} and $\beta$ is the query alignment factor (default $\beta$ as 2.0, with ablations in \S\ref{sec:analysis_2}). 
The key effect of $\alpha_i(t)$ is a training curriculum: early in training, when $\phi(t)$ is small, the query alignment factor is weak and the model is free to explore diverse reasoning languages.
As training progresses and $\phi(t)$ grows, correctly aligned rollouts receive increasingly amplified rewards, gradually guiding the model to converge on the query language.

We specifically adopt Dr.GRPO variant \citep{liu2025understandingr1zeroliketrainingcritical} of GRPO, which removes the length-normalization and standard deviation terms; we find this yields higher reasoning accuracy than standard GRPO (Appendix~\ref{appendix:dr_grpo}). 
For each query $q$, we compute the group advantage $\hat{A}_{i,t}$ over the adjusted rewards $r'=\{r'_1,r'_2,...,r'_\mathcal{O}\}$ as:
\begin{equation}
    \hat{A}_{i,t}=r'_i - \mathrm{mean}(\{r'_1,r'_2,...,r'_\mathcal{O}\}).
\end{equation}
Finally, we optimize the policy model by maximizing the following objective:
\begin{equation}
\begin{split}
    \mathcal{J}(\theta) = \mathbbm{E}_{q\sim P(Q),\{o_i\}^\mathcal{O}_{i=1}\sim {\pi_\theta}_\mathrm{old}(\mathcal{O}|q)} \\ 
    \frac{1}{\mathcal{O}} \sum^\mathcal{O}_{i=1} \sum^{|o_i|}_{t=1} \bigg\{ \mathrm{min} \bigg[ \frac{\pi_\theta (o_{i,t}|q,o_{i,<t})}{{\pi_\theta}_\mathrm{old} (o_{i,t}|q,o_{i,<t})}\hat{A}_{i,t}, \\
    \mathrm{clip} \bigg( \frac{\pi_\theta (o_{i,t}|q,o_{i,<t})}{{\pi_\theta}_\mathrm{old} (o_{i,t}|q,o_{i,<t})}, 1-\epsilon,1+\epsilon \bigg)\hat{A}_{i,t}  \bigg] \\
    - \gamma \cdot \mathrm{KL} [\pi_\theta \Vert \pi_\mathrm{ref}] \bigg\},
\end{split}
\end{equation}
where $\pi_{\mathrm{old}}$ is the frozen old policy and $\pi_\mathrm{ref}$ is the reference model. 
% Further details on the training strategies for both stages are in Appendix~\ref{appendix:train_details}.

%% file: page/03_experimental_setup.tex
\section{\raisebox{-0.2em}{\includegraphics[height=1.1em]{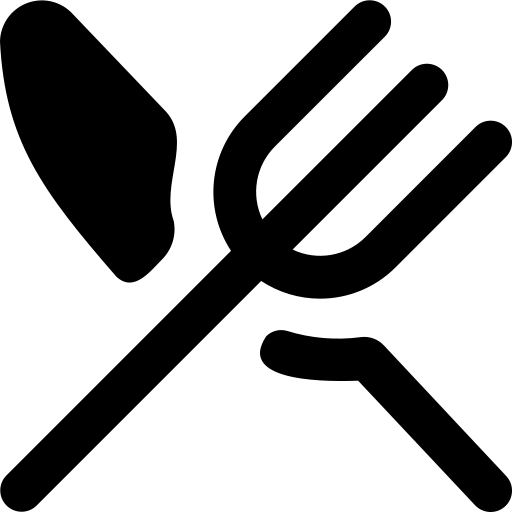}} Experiment Setup}
\label{sec:setup}

\paragraph{Model.}
To assess the effectiveness of AdaMame across models and sizes, we use: \textsc{Distill-Qwen 1.5b} \citep{deepseekai2025deepseekr1incentivizingreasoningcapability} and \textsc{Qwen3 4b} \citep{yang2025qwen3technicalreport} as backbones.
\textsc{Distill-Qwen 1.5b} is built on \textsc{Qwen2.5} and post-trained on multilingual reasoning traces distilled from \textsc{DeepSeek-R1}. 
Further details are in Appendix Table~\ref{tab:model_details}.

% , and \textsc{SmolLM3 3b} \citep{bakouch2025smollm3}. Model details are provided in Appendix~\autoref{tab:model_details}.

\paragraph{Evaluation Dataset.}
We use two multilingual mathematical reasoning benchmarks: MGSM-Rev2 \citep{peter2025mindgapnottranslation} and MSVAMP \citep{chen-etal-2024-breaking}.
MGSM-Rev2 is a revised version of MGSM \citep{shi2022language} with human-translated math problems, which corrects translation errors or ambiguities in 15.8\% queries on average (see Appendix~\ref{appendix:mgsm_query}).
MSVAMP extends SVAMP \citep{patel2021nlp} with \textsc{ChatGPT}-translated problems, with translation quality verified by native speakers.
% PolyMath provides \textsc{GPT-4o}-translated problems across four difficulty tiers (low, medium, high, top).
Detailed dataset statistics are in Appendix Table~\ref{tab:dataset_details}.

\paragraph{Languages.}
We study 12 languages representing diverse resource levels, language families, writing scripts, and linguistic typologies. We select French (fr), Portuguese (pt), Japanese (ja), Korean (ko), and Thai (th) as the in-domain (i.e., training) languages and Bengali (bn), English (en), Spanish (es), Russian (ru), Swahili (sw), Telugu (te), Chinese (zh), and German (de) as out-of-domain languages to measure generalizability. Per-language characteristics are detailed in Appendix Table~\ref{tab:lang_details}.

\input{tables/main_results}

\paragraph{Baselines.}
For each backbone model, we study five baselines of increasing training effort:
\begin{itemize}[leftmargin=*, itemsep=2pt, parsep=-1pt]
    \item \textbf{Vanilla}: The unmodified backbone model $\mathcal{M}$.
    \item \textbf{Prompt}: $\mathcal{M}$ with a language-specific prompt instruction prepended to each query (Appendix~\ref{appendix:prompt}).
    \item \textbf{SFT}: $\mathcal{M}_\mathrm{sft}$, cold-start fine-tuned on $\mathcal{D}_\mathrm{sft}$.
    \item \textbf{GRPO}: $\mathcal{M}_\mathrm{grpo}$, initialized from $\mathcal{M}_\mathrm{sft}$ and trained on $\mathcal{D}_\mathrm{grpo}$ with standard GRPO.
    \item \textbf{AdaMame}: $\mathcal{M}_\mathrm{grpo}$, trained identically but with AdaMame-GRPO instead of standard GRPO.
\end{itemize}
For \textsc{Distill-Qwen 1.5B} backbone model, we additionally compare \textbf{M-Thinker Iter1} and \textbf{Iter2} \citep{zhang2026thinknativelyunlockingmultilingual}, using the checkpoints corresponding to one and two rounds of iterative GRPO training respectively (see Table~\ref{tab:comparison} for details).

\paragraph{Evaluation Metrics.}
We report three metrics:
\begin{itemize}[leftmargin=*, itemsep=2pt, parsep=-1pt]
    \item \textbf{Accuracy (↑)}: Pass@4 over four sampled outputs, which reduces bias and variance associated with single-generation evaluation \citep{zhang2025survey}. Answers are extracted from \texttt{\textbackslash\textbackslash boxed\{\}} using \textsc{math-verify}.\footnote{\href{https://github.com/huggingface/Math-Verify}{\texttt{huggingface/Math-Verify}}}
    
    \item \textbf{LCPR (↑)}: Language Confusion Pass Rate \citep{marchisio-etal-2024-understanding}, computed as the harmonic mean of line- and word-level language fidelity over the reasoning trace $c$.\footnote{Unlike prior work, which measures language fidelity by the top-1 detected language, LCPR explicitly captures undesirable code-switching behavior within traces; see Appendix~\ref{appendix:cs}.}

    \item \textbf{TTC (↓)}: Test-Time Compute, measured as the fraction of the maximum context length (8,192 tokens) consumed by the model's response, reflecting reasoning token efficiency \citep{snell2024scalingllmtesttimecompute, muennighoff-etal-2025-s1}.
\end{itemize}
All inference use \textsc{vLLM} \citep{vllm} with sampling temperature 0.6 and top-p 0.95. Further implementation details are in Appendix~\ref{appendix:metric_details}.

\paragraph{Training Strategies.}
For SFT stage, we adopt LoRA \cite{hu2021loralowrankadaptationlarge} for parameter efficient fine-tuning targeting all linear modules, implemented using \textsc{LLaMA-Factory} \citep{zheng-etal-2024-llamafactory}. 
We set LoRA rank as 8, $\alpha$ as 16, cutoff length as 8,192 tokens, learning rate as $2\mathrm{e}^{-4}$ with cosine scheduler and warm-up ratio 0.1.
% We use \texttt{deepseekr1} template for \textsc{Distill-Qwen 1.5b} and \texttt{qwen3\_think} for \textsc{Qwen-3 4b}
% , and \texttt{smollm3\_think} for \textsc{SmolLM3 3b}.
One SFT run requires 1.5 hours on one NVIDIA-A5000.

For RL stage, we implement all experiments in \textsc{verl} \citep{verl}.
% , with modifications to support the AdaMame-GRPO variant.
We set batch size to 512, mini-batch size to 256 (2 gradient updates per batch), rollout count to 8, max prompt length to 512, max response length to 4,096, training epochs to 10, temperature to 0.8, learning rate $1\mathrm{e}^{-5}$, and KL loss coefficient to 0.001.
The last three hyperparameters are selected via \textsc{wandb}'s sweep search.\footnote{\href{https://docs.wandb.ai/models/ref/python/functions/sweep}{\texttt{wandb.ai/functions/sweep}}}
A single run takes approximately 4 hours on one NVIDIA-A100 for \textsc{Distill-Qwen 1.5b}, 
and 10 hours on two NVIDIA-A100s for \textsc{Qwen3 4b}. 

%% file: tables/main_results.tex
\definecolor{pos}{RGB}{36, 145, 58}
\definecolor{neg}{RGB}{214, 64, 64}

\begin{table*}
\centering
\resizebox{\linewidth}{!}{%
    \large
    \begin{tabular}{ll *{18}{r}}
    \toprule
    \multirow{4}{*}{\textbf{Model}} & \multirow{4}{*}{\textbf{Size}} & \multicolumn{6}{c}{\textbf{Accuracy (\%, ↑)}} &
    \multicolumn{6}{c}{\textbf{Language Fidelity (\%, ↑)}} & \multicolumn{6}{c}{\textbf{Test-Time Compute (\%, ↓)}} \\
    \cmidrule(lr){3-8} \cmidrule(lr){9-14} \cmidrule(lr){15-20}
    & & $\text{G}_\text{in}$ & $\text{G}_\text{out}$ & $\text{G}_\text{all}$ & $\text{V}_\text{in}$ & $\text{V}_\text{out}$ & $\text{V}_\text{all}$ & $\text{G}_\text{in}$ & $\text{G}_\text{out}$ & $\text{G}_\text{all}$ & $\text{V}_\text{in}$ & $\text{V}_\text{out}$ & $\text{V}_\text{all}$ & $\text{G}_\text{in}$ & $\text{G}_\text{out}$ & $\text{G}_\text{all}$ & $\text{V}_\text{in}$ & $\text{V}_\text{out}$ & $\text{V}_\text{all}$ \\
    
    \midrule
    \rowcolor{gray!20}
    \multicolumn{20}{c}{\textbf{\textsc{Distill-Qwen 1.5b}}} \\
    \midrule

    Vanilla & - & 46.9 & 52.8 & 50.8 & 66.5 & 72.0 & 70.4 & 51.9 & 60.3 & 57.5 & 64.1 & \underline{50.2} & 54.4 & 24.7 & 18.4 & 20.5 & 9.7 & 6.3 & 7.3 \\
    Prompt & - & 51.6 & 57.8 & 55.7 & 66.9 & 73.4 & 71.4 & 57.8 & 58.6 & 58.3 & 66.2 & \textbf{52.1} & 56.3 & 29.4 & 34.7 & 33.0 & 10.7 & 6.1 & 7.5 \\
    \textit{M-Thinker Iter1} & 35K & 62.0 & 64.2 & 63.5 & \underline{78.9} & 72.7 & 74.6 & 0.3 & 19.0 & 12.8 & 3.12 & 29.6 & 21.7 & 25.0 & 23.3 & 23.9 & 17.3 & 14.2 & 15.2 \\
    \textit{M-Thinker Iter2} & 50K & 69.5 & \underline{65.0} & 66.5 & \textbf{81.1} & 74.6 & \underline{76.6} & 0.1 & 17.7 & 11.8 & 1.6 & 27.5 & 19.7 & 20.9 & 21.1 & 21.0 & 14.7 & 14.0 & 15.2 \\
    SFT & 30K & 65.3 & 58.3 & 60.6 & 77.5 & 75.4 & 76.0 & 77.6 & \underline{62.8} & 67.7 & 84.0 & 50.1 & \underline{60.3} & 1.5 & 4.2 & \underline{3.3} & \underline{1.2} & \underline{1.7} & \underline{1.5} \\
    +GRPO & 35K & \underline{72.6} & 63.6 & \underline{66.6} & 77.3 & 76.2 & \underline{76.6} & \underline{80.1} & 61.6 & \underline{67.8} & \underline{84.8} & 47.4 & 58.7 & \textbf{1.0} & \underline{2.6} & \textbf{1.9} & \textbf{1.0} & \textbf{1.6} & \textbf{1.4} \\
    +AdaMame-GRPO & 35K & \textbf{72.9} & \textbf{65.5} & \textbf{67.9} & 77.5 & \textbf{76.8} & \textbf{77.0} & \textbf{80.6} & \textbf{64.8} & \textbf{70.1} & \textbf{85.2} & 50.1 & \textbf{60.7} & \underline{1.2} & \textbf{2.3} & \textbf{1.9} & \textbf{1.0} & \textbf{1.6} & \textbf{1.4} \\
    
    $\Delta$ & & \cellcolor{pos!15}{\normalsize{+0.3}} & \cellcolor{pos!35}{\normalsize{+1.9}} & \cellcolor{pos!35}{\normalsize{+1.3}} & \cellcolor{pos!20}{\normalsize{+0.8}} & \cellcolor{pos!15}{\normalsize{+0.4}} & \cellcolor{pos!15}{\normalsize{+0.4}} & \cellcolor{pos!15}{\normalsize{+0.5}} & \cellcolor{pos!50}{\normalsize{+3.2}} & \cellcolor{pos!50}{\normalsize{+2.3}} & \cellcolor{pos!15}{\normalsize{+0.4}} & \cellcolor{pos!50}{\normalsize{+2.7}} & \cellcolor{pos!50}{\normalsize{+2.0}} & \cellcolor{pos!15}{\normalsize{+0.2}} & \cellcolor{pos!15}{\normalsize{-0.1}} & \cellcolor{gray!20}{\normalsize{0.0}} & \cellcolor{gray!20}{\normalsize{0.0}} & \cellcolor{gray!20}{\normalsize{0.0}} & \cellcolor{gray!20}{\normalsize{0.0}} \\
    
    \midrule
    \rowcolor{gray!20}
    \multicolumn{20}{c}{\textbf{\textsc{Qwen3 4b}}} \\
    \midrule

    Vanilla & - & 81.3 & 77.4 & 78.7 & 70.3 & 64.7 & 67.0 & 13.0 & 34.5 & 23.4 & 0.4 & 46.3 & 27.9 & 14.3 & 17.8 & 16.6 & 12.1 & 13.8 & 13.1 \\
    Prompt & - & 88.0 & 80.4 & 82.9 & 77.9 & 82.1 & 80.4 & 14.0 & 36.0 & 24.5 & 0.4 & 46.6 & 28.1 & 14.0 & 17.6 & 16.4 & 13.0 & 14.1 & 13.6 \\
    SFT & 30K & \underline{89.8} & \textbf{83.6} & 85.6 & 86.9 & 88.7 & 88.0 & 87.9 & 88.9 & 88.6 & 93.2 & 78.4 & 84.3 & 2.6 & 1.8 & 1.8 & 1.1 & 1.3 & 1.2 \\
    +GRPO & 35K & 88.9 & \underline{82.9} & \underline{84.9} & \underline{88.0} & \underline{89.0} & \underline{88.6} & \underline{88.3} & \underline{90.3} & \underline{89.6} & \underline{93.7} & \underline{79.2} & \underline{85.0} & \underline{2.1} & \underline{1.6} & \underline{1.6} & \underline{0.8} & \underline{1.1} & \underline{1.0} \\
    +AdaMame-GRPO & 35K & \textbf{90.4} & \textbf{83.6} & \textbf{85.9} & \textbf{88.4} & \textbf{89.4} & \textbf{89.0} & \textbf{88.9} & \textbf{90.5} & \textbf{89.9} & \textbf{94.2} & \textbf{80.4} & \textbf{86.0} & \textbf{1.9} & \textbf{1.4} & \textbf{1.5} & \textbf{0.7} & \textbf{1.0} & \textbf{0.9} \\
     
     \(\Delta\) & & \cellcolor{pos!50}{\normalsize{+1.5}} & \cellcolor{pos!35}{\normalsize{+0.7}} & \cellcolor{pos!50}{\normalsize{+1.0}} & \cellcolor{pos!15}{\normalsize{+0.4}} & \cellcolor{pos!15}{\normalsize{+0.4}} & \cellcolor{pos!15}{\normalsize{+0.4}} & \cellcolor{pos!35}{\normalsize{+0.6}} & \cellcolor{pos!15}{\normalsize{+0.2}} & \cellcolor{pos!15}{\normalsize{+0.3}} & \cellcolor{pos!35}{\normalsize{+0.5}} & \cellcolor{pos!50}{\normalsize{+1.2}} & \cellcolor{pos!50}{\normalsize{+1.0}} & \cellcolor{pos!15}{\normalsize{+0.2}} & \cellcolor{pos!15}{\normalsize{-0.2}} & \cellcolor{pos!15}{\normalsize{-0.1}} & \cellcolor{pos!15}{\normalsize{-0.1}} & \cellcolor{pos!15}{\normalsize{-0.1}} & \cellcolor{pos!15}{\normalsize{-0.1}} \\

    \bottomrule
\end{tabular}
}
\caption{\textbf{Performance across model variants and evaluation datasets.} \textbf{G}: MGSM-Rev2; \textbf{V}: MSVAMP.
\textbf{in}: in-domain; \textbf{out}: out-of-domain languages; \textbf{all}: overall average.
\textbf{Bold} and \underline{underlined} values denote the best and second-best scores per column, respectively.
\(\Delta\) denotes the difference between AdaMame-GRPO and GRPO (Adamame-GRPO$-$GRPO), with darker shading indicating larger magnitude within each model.
AdaMame-GRPO achieves the highest overall reasoning accuracy and LCPR (\textbf{Language Fidelity}) and the lowest TTC (\textbf{Test-Time Compute}) across all variants, using a dataset of 35K instances, which is smaller than or comparable to the \textit{M-Thinker} baselines.
Per-language results are detailed in Appendix~\ref{appendix:per_lang}.}
\label{tab:main_results}
\end{table*}

%% file: page/04_results.tex
\section{\raisebox{-0.2em}{\includegraphics[height=1.5em]{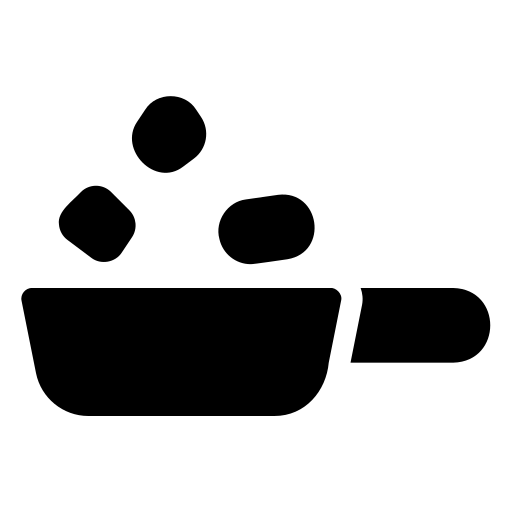}} Results}

We first compare accuracy, LCPR, and TTC across models, in- and out-of-domain languages, and language resource levels (\S\ref{sec:results}).
We then examine adaptivity of AdaMame-GRPO (\S\ref{sec:analysis_1}), and ablate the effect of the query alignment factor $\beta$ (\S\ref{sec:analysis_2}).

\subsection{Main Results}
\label{sec:results}

We present results in Table~\ref{tab:main_results} and summarize several interesting findings below:

\paragraph{Current LRMs struggle with language collapse.}
We observe that both backbone models have low LCPR, indicating that they either frequently default to produce reasoning traces in specific languages regardless of the query language, or to heavily code-switch languages mid-trace.
For example, \textsc{Qwen3 4b} achieves only 23.4\% and 27.9\% LCPR across languages on MGSM-Rev2 and MSVAMP, meaning that nearly one fourth of all cases, the model fails to deliver traces fully in the query language.
This is less pronounced in \textsc{Distill-Qwen 1.5b}, which achieves 54--57\% LCPR across both datasets, likely explained by its explicit post-training on multilingual reasoning traces \citep{deepseekai2025deepseekr1incentivizingreasoningcapability}.

Even with the Prompt baseline, which prepends language-specific instructions to encourage reasoning in the query language, yields only marginal improvements in LCPR (+0.8\%, +1.9\% for \textsc{Distill-Qwen 1.5b}; +1.1\%, +0.2\% for \textsc{Qwen3 4b}, on MGSM-Rev2 and MSVAMP respectively).
This aligns with prior findings that prompt-level interventions are insufficient to fully shift models' reasoning language \citep{qi-etal-2025-models, gao2026explangimprovedexplorationexploitation}.

\input{tables/resource_results}

\paragraph{SFT substantially reduces language collapse but generalizes poorly to out-of-domain languages.}
Across both backbone models and datasets, we show that SFT yields large LCPR gains over the Prompt baseline. This demonstrates that fine-tuning on language-specific reasoning traces is far more effective than lightweight prompt-level intervention at matching the reasoning language to the query \citep{lai-nissim-2024-mcot}.
The gains are considerably larger for \textsc{Qwen3 4b} than \textsc{Distill-Qwen 1.5b} (MGSM-Rev2 / MSVAMP: +9.4\% / +4.0\% for \textsc{Distill-Qwen 1.5b}; +64.1\% / +56.2\% for \textsc{Qwen3 4b}).
Notably, gains in LCPR also translate to improvements in both reasoning accuracy and TTC.
While prior work similarly incorporates SFT as a cold-start step \citep{liu2026selfimprovingmultilinguallongreasoning}, its standalone contribution is unclear and rather has shown to not necessarily improve reasoning accuracy \citep{hwang2025learngloballyspeaklocally}.
Our SFT stage avoids this trade-off, which we attribute to the use of non-MT reasoning traces in $\mathcal{D}_\text{sft}$: unlike machine-translated traces, these are both fluent in the query language and reflect language-specific reasoning patterns, providing a higher-quality training signal for multilingual reasoning (details in Appendix~\ref{appendix:naturally}).

\input{figures/lang_match}

However, SFT alone generalizes poorly beyond its five training languages: on out-of-domain languages, LCPR gains are substantially smaller and sometimes fall below the Prompt baseline (e.g., -2.0\% for \textsc{Distill-Qwen 1.5b} on MSVAMP).

\paragraph{GRPO improves out-of-domain reasoning accuracy; AdaMame-GRPO further improves LCPR generalization.}

We show that GRPO yields stronger accuracy gains on out-of-domain languages than SFT.
For instance, \textsc{Distill-Qwen 1.5b} gains +0.5\% accuracy on out-of-domain languages over SFT, but an additional +5.3\% via GRPO.
This is consistent with the established roles of the two training paradigms: SFT encourages memorization of patterns seen during training \citep{allen2023physics, kang2024learning}, while RL encourages learning generalizable rules that transfer to new settings \citep{chu2025sft}.
However, GRPO yields comparatively smaller LCPR gains than accuracy gains (+0.1\% vs. +6.0\% on MGSM-Rev2 for \textsc{Distill-Qwen 1.5b}), as the accuracy-only reward of GRPO provides no incentive to align the reasoning language with the query.

AdaMame-GRPO addresses this by introducing the query alignment factor as a scaling term on the accuracy reward (\S\ref{sec:method_2}), achieving higher LCPR than GRPO without sacrificing, but instead further improving, reasoning accuracy.
This stands in contrast to the accuracy-language fidelity trade-off observed in prior work \citep{wang2025polymath}.
AdaMame-GRPO with \textsc{Distill-Qwen 1.5b} also outperforms \textit{M-Thinker Iter2} baseline, which uses 15K more training instances (35K vs. 50K), showing greater data efficiency.
Specifically, AdaMame-GRPO shows a substantial LCPR advantage over \textit{M-Thinker} baselines, which we attribute to heavy code-switching in \textit{M-Thinker}'s reasoning traces (Appendix~\ref{appendix:code-switch}). It also reflects a failure mode that the top-1 language detection metric in prior work overlooks but LCPR explicitly penalizes.

Taken together, we show that AdaMame-GRPO resolves the key limitations identified across prior approaches (Figure~\ref{fig:teaser_fig}):
poor out-of-domain generalization from SFT, limited LCPR gains from standard GRPO, and overthinking that persist in reward-concatenation methods \citep{zhang2026thinknativelyunlockingmultilingual}, by achieving the lowest TTC.\footnote{We compare the length distribution of the SFT training data against the reasoning traces produced by each baseline in Appendix~\ref{appendix:length}, and show that AdaMame-GRPO-generated traces are the shortest on average.}

\paragraph{AdaMame-GRPO yields the largest gains on lower-resource languages.}
Table~\ref{tab:resource_results} reports performance broken down by language resource level (low, mid, high) to examine whether AdaMame-GRPO's gains are consistent across different languages.
For \textsc{Distill-Qwen 1.5b}, while \textit{M-Thinker} baselines tend to lead on high-resource languages, AdaMame-GRPO shows a clear advantage on low-resource languages (Thai, Bengali, Swahili, and Telugu), which is notable given that only Thai appears in the SFT and RL training corpora.
For example, on MSVAMP, it improves accuracy by +10.8\% for Bengali on \textsc{Distill-Qwen 1.5b} and +28.4\% on \textsc{Qwen3 4b} over the Vanilla baseline.

The same pattern holds for LCPR: AdaMame-GRPO achieves the highest low-resource LCPR on both datasets, while Vanilla and Prompt baselines retain an advantage only at the high-resource tier.
This positions our approach as a step toward more equitable multilingual reasoning~\citep{tran2025reasoningtransferextremelylowresource, zhao2026when}, where both accuracy and LCPR gains are not only in well-resourced languages at the expense of underrepresented ones.

\subsection{Adaptivity of AdaMame-GRPO}
\label{sec:analysis_1}

% In order to verify that the query alignment factor in AdaMame-GRPO (\S\ref{sec:method_2}) indeed makes the model to progressively \textit{adapt} its reasoning language to the query language, we examine how the reasoning language distribution evolves over the course of training.
% As shown in~\autoref{fig:lang_match}, both models follow a natural curriculum: they initially explore reasoning languages other than the query, then gradually converge toward it, with an overall upward trend in language adaptation rate.

We verify that the cosine growth schedule in AdaMame-GRPO induces the intended explore-then-exploit curriculum: as shown in Figure~\ref{fig:lang_match}, models initially generate rollouts across diverse reasoning languages, then progressively shift toward the query language as the alignment factor grows.

\definecolor{fr}{HTML}{FFBD00}
\definecolor{val}{HTML}{6F5849}
\definecolor{pt}{HTML}{81B29A}
\definecolor{ja}{HTML}{3D405B}
\definecolor{ko}{HTML}{E07A5F}
\definecolor{th}{HTML}{CE4257}

Interestingly, we find distinct per-language trajectories, which suggest that AdaMame-GRPO adaptively rebalances the accuracy and language alignment objectives at the language level rather than uniformly.
While \textcolor{fr}{\ding{108}} French shows a monotonically increasing trend that plateaus after roughly the first quarter of training, the remaining four languages (\textcolor{pt}{\ding{117}}, \textcolor{ja}{\ding{110}}, \textcolor{ko}{\ding{115}}, \textcolor{th}{\ding{116}}) each exhibit a mid-training dip around the halfway point before recovering to a higher endpoint. 
This dip likely reflects an exploration vs. exploitation trade-off: as the accuracy reward still dominates early in training, the model temporarily shifts toward higher-resource languages that yield more correct rollouts, sacrificing language alignment before the growing alignment factor re-balances the two objectives.
Consistent with this interpretation, \textcolor{val}{\ding{72}} validation accuracy rises relatively steadily through the dip, suggesting the model is exploiting accurate but language-misaligned reasoning before converging on both.
% For \textsc{Qwen3 4B}, the dip is accompanied by a simultaneous drop in validation accuracy around batch 4, indicating a more disruptive transition phase, likely due to its lack of explicit multilingual reasoning post-training.

\textsc{Distill-Qwen 1.5b} also shows larger final gains on lower-resource languages (\textcolor{ja}{\ding{110}} Japanese, \textcolor{ko}{\ding{115}} Korean, and \textcolor{th}{\ding{116}} Thai), while \textsc{Qwen3 4b} shows a relative advantage on higher-resource languages (\textcolor{fr}{\ding{108}} French and \textcolor{pt}{\ding{117}} Portuguese), consistent with the resource-level breakdown results in Table~\ref{tab:resource_results}.

\input{figures/change_in_b_vis}

\subsection{Ablation on Query Alignment Factor}
\label{sec:analysis_2}

\definecolor{mgsm}{HTML}{9E2F5B}
\definecolor{msvamp}{HTML}{2A9A92}

While we set the query alignment factor $\beta$ to 2.0 as the default in AdaMame-GRPO (\S\ref{sec:method_2}), we analyze its sensitivity by varying $\beta$ across a range of values.
Figure~\ref{fig:change_in_b} reports reasoning accuracy and LCPR across both backbone models and datasets: \textcolor{mgsm}{\ding{108}}/\textcolor{mgsm}{\ding{110}} MGSM-Rev2 and \textcolor{msvamp}{\ding{108}}/\textcolor{msvamp}{\ding{110}} MSVAMP.

We observe that increasing $\beta$ consistently improves LCPR but at the cost of reasoning accuracy, showing a controllable accuracy-language fidelity trade-off relation.
% We find $\beta$ as 2.0 to be the optimal setting and adopt it in all main experiments.
Between the two models, \textsc{Qwen3 4b} is more sensitive to changes in $\beta$ than \textsc{Distill-Qwen 1.5b} on both metrics, likely because it was not explicitly post-trained on multilingual reasoning traces.
The trade-off observed here confirms two properties of the query alignment factor: it reliably improves language fidelity to the query language as intended, but at the same time, it introduces a conflicting signal with the accuracy objective, which can be controlled through $\beta$.

%% file: tables/resource_results.tex
\begin{table*}
\centering
\resizebox{\linewidth}{!}{%
    \large
    \begin{tabular}{ll *{18}{r}}
    \toprule
    \multirow{4}{*}{\textbf{Model}} & \multirow{4}{*}{\textbf{Size}} & \multicolumn{6}{c}{\textbf{Accuracy (\%, ↑)}} &
    \multicolumn{6}{c}{\textbf{Language Fidelity (\%, ↑)}} & \multicolumn{6}{c}{\textbf{Test-Time Compute (\%, ↓)}} \\
    \cmidrule(lr){3-8} \cmidrule(lr){9-14} \cmidrule(lr){15-20}
    & & $\text{G}_\text{low}$ & $\text{G}_\text{mid}$ & $\text{G}_\text{high}$ & $\text{V}_\text{low}$ & $\text{V}_\text{mid}$ & $\text{V}_\text{high}$ & $\text{G}_\text{low}$ & $\text{G}_\text{mid}$ & $\text{G}_\text{high}$ & $\text{V}_\text{low}$ & $\text{V}_\text{mid}$ & $\text{V}_\text{high}$ & $\text{G}_\text{low}$ & $\text{G}_\text{mid}$ & $\text{G}_\text{high}$ & $\text{V}_\text{low}$ & $\text{V}_\text{mid}$ & $\text{V}_\text{high}$ \\
    
    \midrule
    \rowcolor{gray!20}
    \multicolumn{20}{c}{\textbf{\textsc{Distill-Qwen 1.5b}}} \\
    \midrule

    Vanilla & - & 15.7 & 56.0 & 80.8 & 36.1 & 80.5 & 88.4 & 39.1 & 40.9 & \textbf{92.5} & 33.0 & 21.9 & \underline{94.8} & 40.7 & 12.9 & 7.83 & 12.7 & 6.1 & 4.1 \\
    Prompt & - & 21.5 & 61.9 & 83.7 & 38.8 & 81.2 & 88.6 & 39.7 & 49.3 & 85.9 & 38.5 & 21.9 & \textbf{95.5} & 62.5 & 22.9 & 13.5 & 14.8 & 6.0 & 3.1 \\
    \textit{M-Thinker Iter1} & 35K & 30.6 & 69.5 & \underline{90.3} & 51.0 & 81.1 & 87.3 & 2.8 & 1.5 & 34.0 & 6.6 & 5.5 & 45.0 & 16.9 & 26.2 & 28.5 & 14.2 & 16.0 & 15.2 \\
    \textit{M-Thinker Iter2} & 50K & 34.4 & \underline{74.7} & \textbf{90.4} & 51.9 & \underline{83.3} & \textbf{90.0} & 2.3 & 0.2 & 33.0 & 4.3 & 3.0 & 43.9 & 17.2 & 23.2 & 22.7 & 13.9 & 15.1 & 13.7 \\
    SFT & 30K & 31.4 & 66.9 & 83.5 & 52.4 & 81.5 & \underline{89.6} & \underline{55.4} & 59.8 & \underline{88.1} & \underline{49.1} & \underline{31.9} & 89.9 & 6.4 & 1.9 & \underline{1.7} & \underline{2.6} & \underline{1.1} & 1.0 \\
    +GRPO & 35K & \underline{38.5} & 73.8 & 87.4 & \underline{53.2} & 82.0 & \textbf{90.0} & 55.0 & \textbf{61.2} & 87.1 & 40.8 & \textbf{34.6} & 90.0 & \underline{3.5} & \textbf{1.1} & \textbf{1.2} & \underline{2.6} & \underline{1.1} & \textbf{0.8} \\
    +AdaMame-GRPO & 35K & \textbf{39.5} & \textbf{74.9} & 89.4 & \textbf{53.6} & \textbf{83.8} & 89.5 & \textbf{61.4} & \underline{61.0} & 87.8 & \textbf{49.8} & 31.7 & 90.5 & \textbf{3.2} & \underline{1.4} & \textbf{1.2} & \textbf{2.5} & \textbf{0.9} & \underline{0.9} \\

    \midrule
    \rowcolor{gray!20}
    \multicolumn{20}{c}{\textbf{\textsc{Qwen3 4b}}} \\
    \midrule

    Vanilla & - & \textbf{74.1} & 79.5 & 82.4 & 64.9 & 64.5 & 70.3 & 4.1 & 23.5 & 42.7 & 0.6 & 30.7 & 46.4 & 21.6 & 13.9 & 14.3 & 17.2 & 11.4 & 11.2 \\
    Prompt & - & 69.0 & 88.4 & 91.3 & 70.6 & 78.5 & 89.2 & 6.0 & 23.7 & 43.6 & 1.2 & 30.8 & 46.2 & 23.0 & 12.7 & 13.6 & 19.1 & 10.6 & 11.8 \\
    SFT & 35K & 70.8 & \textbf{91.1} & 95.0 & 77.6 & 91.5 & \underline{93.1} & 88.7 & 84.9 & \underline{92.2} & \underline{64.1} & 91.9 & 93.8 & 3.4 & \textbf{1.0} & \textbf{1.0} & 1.7 & \underline{0.9} & \underline{1.0} \\
    +GRPO & 35K & 69.4 & 90.1 & \underline{95.1} & \underline{79.5} & \textbf{92.0} & 92.9 & \underline{89.0} & \underline{87.3} & \textbf{92.5} & 63.7 & \underline{93.2} & \underline{94.8} & \underline{2.8} & \underline{1.2} & \underline{1.1} & \underline{1.4} & \underline{0.9} & \textbf{0.8} \\
    +AdaMame-GRPO & 35K & \underline{71.1} & \underline{90.9} & \textbf{95.6} & \textbf{80.0} & \underline{91.9} & \textbf{93.6} & \textbf{89.4} & \textbf{87.8} & \textbf{92.5} & \textbf{64.7} & \textbf{93.8} & \textbf{96.0} & \textbf{2.6} & \textbf{1.0} & \textbf{1.0} & \textbf{1.2} & \textbf{0.8} & \textbf{0.8} \\

    \bottomrule
\end{tabular}
}
\caption{\textbf{Performance by language resource level.} \textbf{G}: MGSM-Rev2; \textbf{V}: MSVAMP.
\textbf{low}: low-resource (\textit{th}, bn, sw, te); \textbf{mid}: mid-resource (\textit{ja}, \textit{ko}, \textit{pt}, ru, de); \textbf{high}: high-resource (\textit{fr}, en, es, zh).
Resource level is determined by number of speakers and Wikipedia article count per language (Appendix~\autoref{tab:lang_details}). 
Note that there is at least one in-domain language (\textit{italic} in the list) in each resource level group.
AdaMame-GRPO shows the greatest gains over baselines on lower-resource languages.}
\label{tab:resource_results}
\end{table*}

%% file: figures/lang_match.tex
\begin{figure*}
    \centering
    % \begin{minipage}[t]{0.49\linewidth}
    %     \centering
    %     \includegraphics[width=\linewidth]{figures/lang_match_distill_acc.png}
    %     % \caption*{(a) \textsc{Distill-Qwen 1.5b}}
    % \end{minipage}
    % \hfill
    % \begin{minipage}[t]{0.49\linewidth}
    %     \centering
    %     \includegraphics[width=\linewidth]{figures/lang_match_qwen_acc.png}
    %     % \caption*{(b) \textsc{Qwen3 4b}}
    % \end{minipage}
    \includegraphics[width=\linewidth]{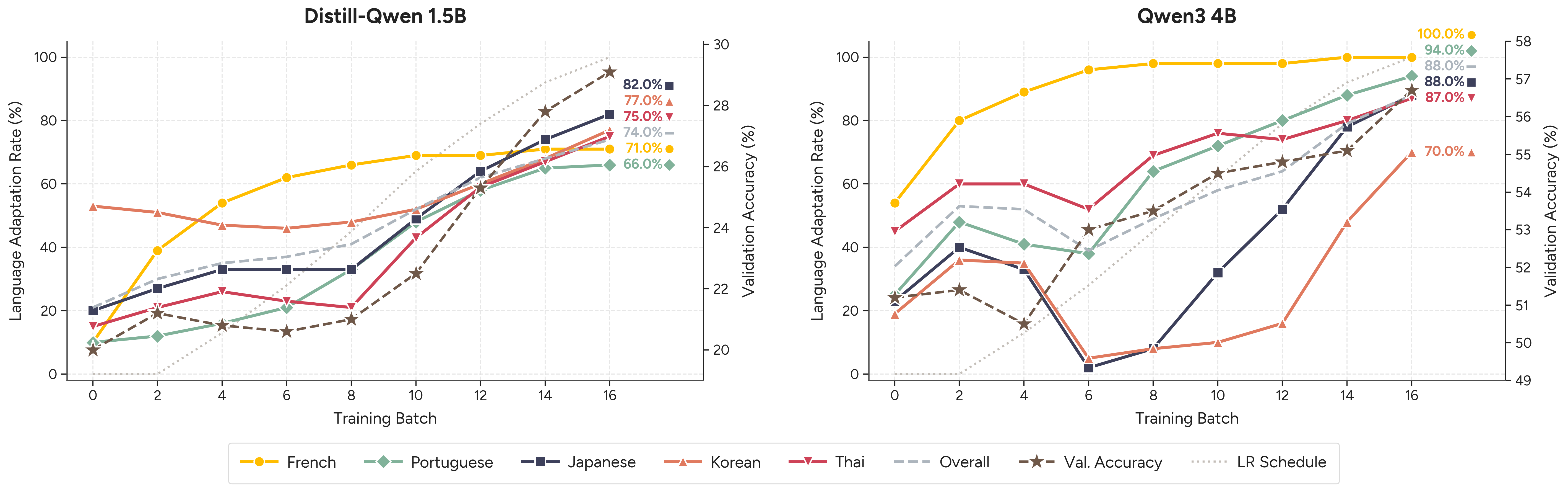}

    \caption{\textbf{Reasoning language adaptation rate and validation accuracy throughout AdaMame-GRPO training.} Left: \textsc{Distill-Qwen 1.5b}; Right: \textsc{Qwen3 4b}.
    The $x$-axis denotes training batch; the left $y$-axis shows the proportion of rollouts whose reasoning trace matches the query language (language adaptation rate; range: 0--1) and the right $y$-axis shows the validation accuracy (\%, \(\uparrow \)). Both are evaluated on a held-out validation set of 1K instances (250 per language) every two training batches.}
    \label{fig:lang_match}
\end{figure*}

%% file: figures/change_in_b_vis.tex
\begin{figure*}
    \centering
    \includegraphics[width=0.95\linewidth]{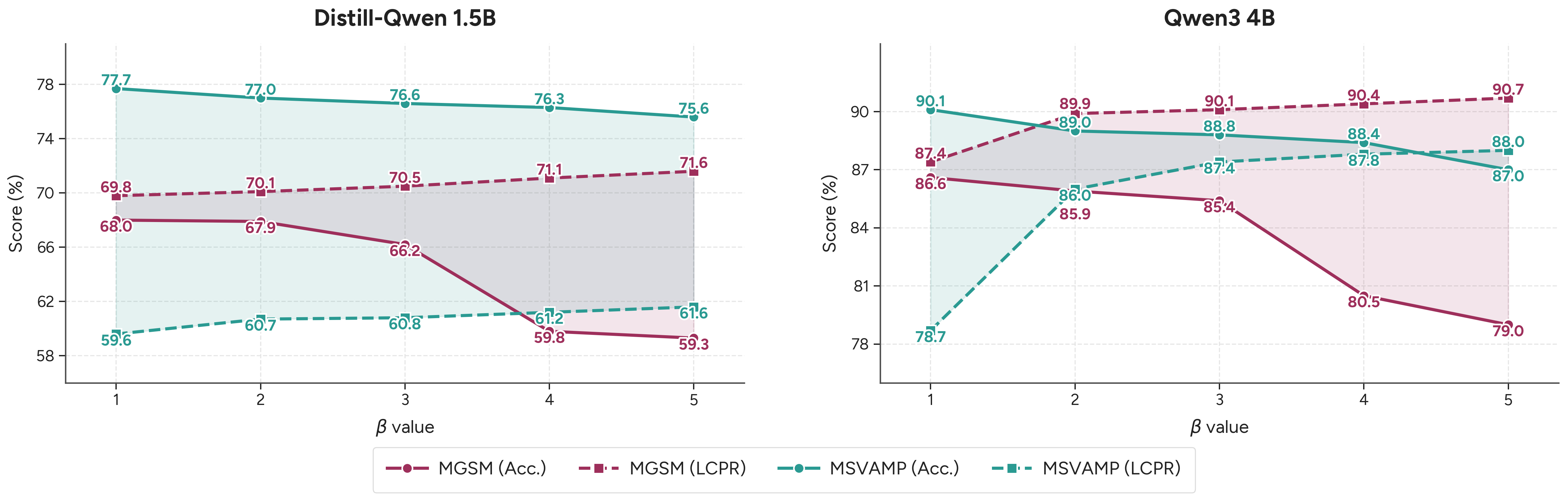}
    \caption{\textbf{Accuracy and LCPR across values of $\beta$.} $\beta$ controls the weight of the query alignment factor in AdaMame-GRPO. Increasing $\beta$ trades-off accuracy for higher LCPR.
    Full numerical results are in Appendix~\ref{appendix:change_in_b}.}
    \label{fig:change_in_b}
\end{figure*}

%% file: page/06_conclusion.tex
\section{Conclusion}

% Original - longer version
% In this work, we present AdaMame, a two-stage training recipe for multilingual mathematical reasoning that jointly improves reasoning accuracy, language fidelity, and token efficiency.
% The SFT stage establishes foundational multilingual reasoning capability from naturally occurring reasoning traces, while AdaMame-GRPO introduces a query-conditioned scaling factor that guides the model to explore diverse reasoning languages early in training before converging on the query language.
% Experiments across two benchmarks, two LRMs, and 12 languages show that AdaMame consistently outperforms all baselines, with the strongest gains on out-of-domain and lower-resource languages.
% Importantly, these gains do not come at the cost of accuracy. Unlike prior reward concatenation approaches, AdaMame resolves the persistent trade-off between reasoning accuracy and language fidelity by treating alignment as a growing objective rather than a fixed competing term.
% We envision AdaMame and the accompanying dataset encourage future work to move beyond English-centric reward designs toward training recipes that are responsive to the full diversity of languages users bring to these systems.

% Shorter version
We present AdaMame, a two-stage training recipe featuring AdaMame-GRPO with a query alignment scaling factor that guides models to explore diverse reasoning languages before exploiting reasoning in the query language.
AdaMame-GRPO achieves Pareto-optimal performance across reasoning accuracy, language fidelity, and token efficiency, with the strongest gains on out-of-domain and lower-resource languages.
We hope this work encourages moving beyond English-centric reward designs toward training recipes that serve the full diversity of languages users bring to these systems.

%% file: page/07_limitation.tex
\section*{Limitations}

\paragraph{Language Detection Reliability.}
The query alignment reward in AdaMame-GRPO relies on the \textsc{lingua} language detector, whose signals may be less reliable for lower-resource languages or heavily code-switched traces. To partially mitigate this, we validate detector reliability on a held-out test set in Appendix~\ref{appendix:reliability}.

\paragraph{Computational Scope.}
Our experiments are constrained by computational budget, which limits both the backbone model sizes and training strategies explored. 
% Among the prior approaches in Table~\ref{tab:comparison}, we compare only against M-Thinker \citep{zhang2026thinknativelyunlockingmultilingual}, as the others have either not released their models or use different backbone models, precluding a fair comparison.
AdaMame is further evaluated only on mathematical reasoning, where multilingual benchmarks are readily available. This leaves open questions on how AdaMame generalizes to other settings, including larger LRMs, non-mathematical tasks, and a broader set of languages, which we leave for future work.
% (1) -> TRIT, ReasonXL
% (2) -> SP3F

% \paragraph{Dependence on \textsc{GPT-5 nano}.}
% Our training corpora $\mathcal{D}_\mathrm{sft}$ and $\mathcal{D}_\mathrm{grpo}$ consist of reasoning traces generated by \textsc{GPT-5 nano}. 
% We select this model based on its highest average retain rate across the five in-domain languages in our preliminary experiments, compared to \textsc{Qwen3 32b} and \textsc{Distill-Qwen 32b}. 
% Nevertheless, the resulting corpora may inherit biases from \textsc{GPT-5 nano}'s language-specific reasoning behaviors.

\paragraph{Generalization of LCPR.}
The current word-level language confusion evaluation metric (Appendix~\ref{appendix:metric_details}) only considers code-switching to English, as originally proposed by \citet{marchisio-etal-2024-understanding}. 
We adopt this formulation directly because both our qualitative analyses and the empirical findings of \citet{lin2026thinkmultilingualharderdataefficient} examining code-switching in multilingual reasoning show that English is the most common reasoning language when models are prompted in non-English languages.
It would be valuable to extend the current word-level metric to capture code-switching into arbitrary languages, in order to improve its robustness. Since this would first require substantial annotation effort to identify erroneous non-English words that do not typically occur in the target-language text, we view this as a promising direction for future work.

%% file: page/09_acknowledgement.tex
\section*{Acknowledgments}

We thank the members of the \textsc{clip} lab at the University of Maryland for their valuable feedback and support, with special thanks to Yekyung Kim and Dang Nguyen for their helpful advice on GRPO training. We also thank Siye Wu for graciously answering our questions about ARM and Sanchit Ahuja for insightful discussions on the earlier version of the work.

%% file: figures/prompt.tex
\begin{figure*}
    \centering
    \begin{mdframed}[
        linecolor=lightgrayframe,
        linewidth=0.8pt,
        roundcorner=2pt,
        backgroundcolor=white
    ]
        \includegraphics[width=\linewidth]{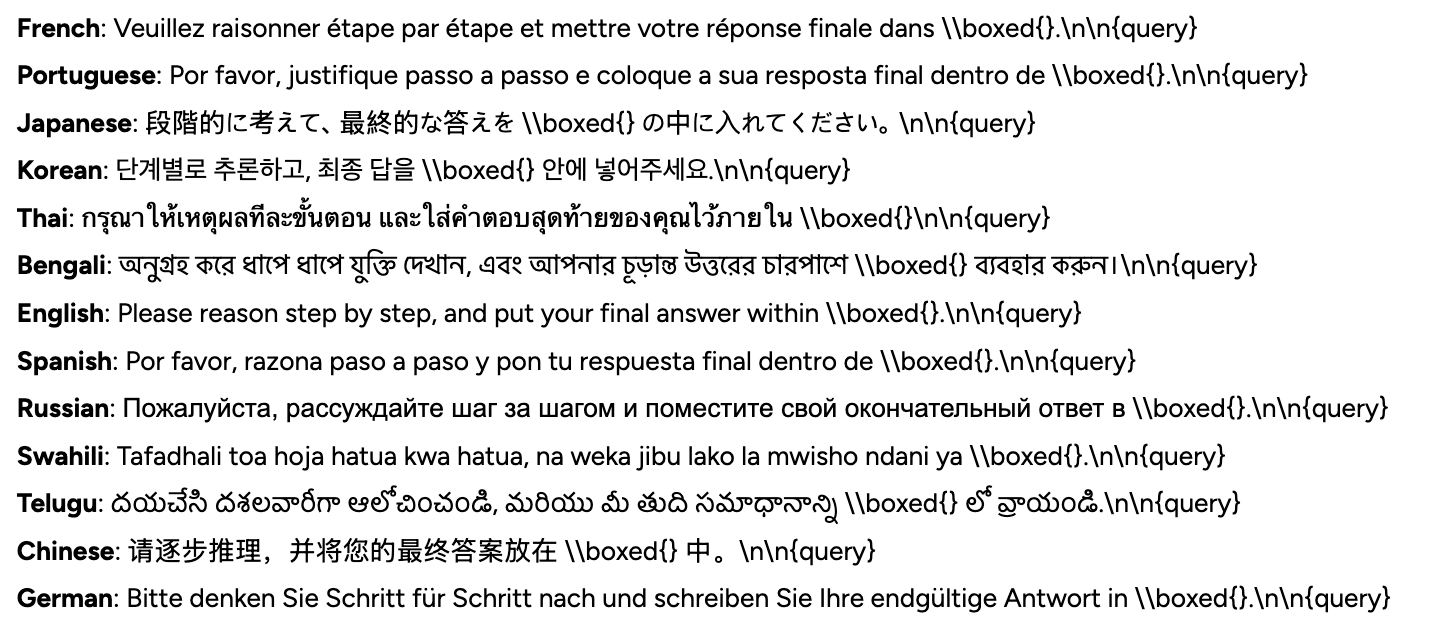}
    \end{mdframed}
    \caption{\textbf{Prompt templates used for sampling generations for each language.}}
    \label{fig:prompt}
\end{figure*}

%% file: tables/naturally.tex
\begin{table}[H]
\centering
\resizebox{\linewidth}{!}{%
    \begin{tabular}{lrr}
    \toprule
    \textbf{\textsc{Distill-Qwen 1.5b}} & \textbf{Accuracy (\%, ↑)} & \textbf{LCPR (\%, ↑)} \\
    \toprule

    \textbf{w/ non-MT} & 60.6 & 67.7 \\
    \textbf{w/ non-MT+MT} & 56.4 & 63.0 \\

    \bottomrule
    \end{tabular}
}
\caption{\textbf{Reasoning accuracy and LCPR for different dataset variants.} Backbone model: \textsc{Distill-Qwen 1.5b}.}  
\label{tab:naturally}
\end{table}

%% file: tables/lora.tex
\begin{table}[H]
\centering
\resizebox{\linewidth}{!}{%
    \begin{tabular}{lrr}
    \toprule
    \textbf{\textsc{Distill-Qwen 1.5b}} & \textbf{Accuracy (\%, ↑)} & \textbf{LCPR (\%, ↑)} \\
    \toprule

    \textbf{w/ LoRA} & 60.6 & 67.7 \\
    \textbf{w/ Full fine-tuning} & 59.2 & 65.4 \\

    \bottomrule
    \end{tabular}
}
\caption{\textbf{Reasoning accuracy and LCPR for LoRA and full fine-tuning.} Backbone model: \textsc{Distill-Qwen 1.5b}.}  
\label{tab:lora}
\end{table}

%% file: tables/format.tex
\begin{table}[H]
\centering
\resizebox{\linewidth}{!}{%
    \begin{tabular}{lrr}
    \toprule
    \textbf{\textsc{Distill-Qwen 1.5b}} & \textbf{Accuracy (\%, ↑)} & \textbf{LCPR (\%, ↑)} \\
    \toprule

    \textbf{Accuracy} & 67.9 & 70.1 \\
    \textbf{Accuracy+Format} & 66.2 & 67.1 \\

    \bottomrule
    \end{tabular}
}
\caption{\textbf{Reasoning accuracy and LCPR for different reward designs.} Backbone model: \textsc{Distill-Qwen 1.5b}.}  
\label{tab:format}
\end{table}

%% file: tables/scheduler.tex
\begin{table}[H]
\centering
\resizebox{\linewidth}{!}{%
    \begin{tabular}{lrr}
    \toprule
    \textbf{\textsc{Distill-Qwen 1.5b}} & \textbf{Accuracy (\%, ↑)} & \textbf{LCPR (\%, ↑)} \\
    \toprule

    \textbf{Cosine growth} & 67.9 & 70.1 \\
    \textbf{Threshold-based} & 66.3 & 68.2 \\
    \textbf{Lagrangian multiplier} & 66.4 & 68.9 \\
    
    \bottomrule
    \end{tabular}
}
\caption{\textbf{Reasoning accuracy and LCPR for scheduler variants.} Backbone model: \textsc{Distill-Qwen 1.5b}.}  
\label{tab:scheduler}
\end{table}

%% file: tables/dr_grpo.tex
\begin{table}[H]
\centering
\resizebox{\linewidth}{!}{%
    \begin{tabular}{lrr}
    \toprule
    \textbf{\textsc{Distill-Qwen 1.5b}} & \textbf{Accuracy (\%, ↑)} & \textbf{LCPR (\%, ↑)} \\
    \toprule

    \textbf{GRPO} & 66.6 & 67.8 \\
    \textbf{Dr.GRPO} & 66.7 & 68.3 \\
    \textbf{AdaMame-GRPO w/ GRPO} & 63.4 & 69.9 \\
    \textbf{AdaMame-GRPO w/ Dr.GRPO} & 67.9 & 70.1 \\

    \bottomrule
    \end{tabular}
}
\caption{\textbf{Reasoning accuracy and LCPR for GRPO and Dr.GRPO.} Backbone model: \textsc{Distill-Qwen 1.5b}.}  
\label{tab:dr_grpo}
\end{table}

%% file: tables/sampling.tex
\begin{table}[H]
\centering
\resizebox{\linewidth}{!}{%
    \begin{tabular}{lrr}
    \toprule
    \textbf{\textsc{Distill-Qwen 1.5b}} & \textbf{Accuracy (\%, ↑)} & \textbf{LCPR (\%, ↑)} \\
    \toprule

    \textbf{Random} & 59.8 & 62.1 \\
    \textbf{Conditional} & 59.7 & 64.6 \\
    \textbf{Rejection} & 67.9 & 70.1 \\

    \bottomrule
    \end{tabular}
}
\caption{\textbf{Reasoning accuracy and LCPR for different sampling strategies.} Backbone model: \textsc{Distill-Qwen 1.5b}.}  
\label{tab:sampling}
\end{table}

%% file: tables/distribution.tex
\begin{table}[H]
\centering
\resizebox{0.8\linewidth}{!}{%
    \begin{tabular}{lrr}
    \toprule
    \textbf{$|\mathcal{O}_\mathrm{correct}|$} & \textbf{Count} & \textbf{Proportion (\%)} \\
    \toprule

    \textbf{1} & 1,342 & 26.8 \\
    \textbf{2} & 1,358 & 27.2 \\
    \textbf{3} & 918 & 18.4 \\
    \textbf{4} & 573 & 11.5 \\
    \textbf{5} & 387 & 7.70 \\
    \textbf{6} & 243 & 4.91 \\
    \textbf{7} & 179 & 3.63 \\

    \bottomrule
    \end{tabular}
}
\caption{\textbf{Distribution of $|\mathcal{O}_\mathrm{correct}|$.}}  
\label{tab:distribution}
\end{table}

%% file: tables/lang_detection.tex
\begin{table}
\centering
\resizebox{0.7\linewidth}{!}{%
    \begin{tabular}{lrr}
    \toprule
    \textbf{Language} & \textbf{Short (\%)} & \textbf{Long (\%)} \\
    \toprule

    \textbf{French} & 100.0 & 100.0 \\
    \textbf{Japanese} & 100.0 & 99.5 \\
    \textbf{Korean}  & 100.0 & 100.0 \\
    \textbf{Thai}  & 100.0 & 99.0 \\
    \textbf{Bengali}  & 100.0 & 98.5 \\
    \textbf{English}  & 100.0 & 100.0 \\
    \textbf{Spanish}  & 100.0 & 99.5 \\
    \textbf{Russian}  & 100.0 & 98.0 \\
    \textbf{Swahili}  & 100.0 & 100.0 \\
    \textbf{Telugu}  & 100.0 & 98.5 \\
    \textbf{Chinese}  & 100.0 & 98.5 \\

    \bottomrule
    \end{tabular}
}
\caption{\textbf{Language detection reliability by language.} \textbf{Short}: short-context evaluation on MGSM-Rev2 queries; \textbf{Long}: long-context evaluation on mCoT-MATH traces.}  
\label{tab:lang_detection}
\end{table}

%% file: tables/filtering.tex
\begin{table}[H]
\centering
\resizebox{0.8\linewidth}{!}{%
    \begin{tabular}{lr}
    \toprule
    \textbf{Language} & \textbf{Retain rate} \\
    \toprule

    \textbf{French} & 6,330 / 8,716 = 72.6\% \\
    \textbf{Japanese} & 6,226 / 8,495 = 73.3\% \\
    \textbf{Korean} & 5,763 / 8,330 = 69.2\% \\
    \textbf{Portuguese} & 6,277 / 8,648 = 72.6\% \\
    \textbf{Thai} & 6,366 / 8,667 = 73.5\% \\

    \bottomrule
    \end{tabular}
}
\caption{\textbf{Retain rate after the filtering process.}}  
\label{tab:filtering}
\end{table}

%% file: tables/stat1.tex
\begin{table*}
\centering
\resizebox{0.9\linewidth}{!}{%
    \begin{tabular}{lrrrrr}
    \toprule
    \textbf{Length bin} & \textbf{French} & \textbf{Japanese} & \textbf{Korean} & \textbf{Portuguese} & \textbf{Thai} \\
    \toprule
    
    $<$500      & 537 (10.0\%)  & 2,209 (36.4\%) & 1,670 (29.5\%) & 305 (5.1\%)    & 529 (8.6\%) \\
    500--750    & 1,323 (24.7\%) & 1,980 (32.7\%) & 1,784 (31.5\%) & 1,017 (16.9\%) & 1,227 (19.9\%) \\
    750--1k     & 1,289 (24.1\%) & 1,188 (19.6\%) & 1,172 (20.7\%) & 1,308 (21.7\%) & 1,475 (23.9\%) \\
    1k--2k      & 2,033 (38.0\%) & 679 (11.2\%)   & 1,026 (18.1\%) & 3,076 (51.1\%) & 2,733 (44.4\%) \\
    2k--4k      & 171 (3.2\%)    & 7 (0.1\%)      & 11 (0.2\%)     & 316 (5.2\%)    & 198 (3.2\%) \\
    4k+         & 2 (0.0\%)      & 0 (0.0\%)      & 0 (0.0\%)      & 2 (0.0\%)      & 0 (0.0\%) \\
    
    \bottomrule
    \end{tabular}
}
\caption{\textbf{Length of multilingual reasoning traces.} Average length per language is: French (989), Japanese (638), Korean (711), Portuguese (1,139), Thai (1,039).}  
\label{tab:stat1}
\end{table*}

%% file: tables/stat2.tex
\begin{table*}
\centering
\resizebox{0.7\linewidth}{!}{%
    \begin{tabular}{lrrrrr}
    \toprule
    \textbf{Metric} & \textbf{French} & \textbf{Japanese} & \textbf{Korean} & \textbf{Portuguese} & \textbf{Thai} \\
    \toprule
    
    Faithfulness (\%, $\uparrow$) & 91.5 & 90.2 & 91.9 & 91.8 & 92.8 \\
    Error rate (\%, $\downarrow$) & 87.3 & 85.5 & 86.1 & 86.4 & 87.7 \\

    \bottomrule
    \end{tabular}
}
\caption{\textbf{Faithfulness and error rate of multilingual reasoning traces.} \textsc{GPT-5.5} as LLM-as-judge.}  
\label{tab:stat2}
\end{table*}

%% file: tables/mgsm_details.tex
\begin{table}[H]
\centering
\resizebox{0.6\linewidth}{!}{%
    \begin{tabular}{lr}
    \toprule
    \textbf{Language} & \textbf{\# Updated (\%)} \\
    \toprule

    \textbf{Bengali} & 46 (18.4\%) \\
    \textbf{English} & 22 (8.80\%) \\
    \textbf{German} & 35 (14.0\%) \\
    \textbf{Spanish} & 38 (15.2\%) \\
    \textbf{French} & 45 (18.0\%) \\
    \textbf{Russian} & 32 (12.8\%) \\
    \textbf{Swahili} & 43 (17.2\%)\\
    \textbf{Telugu} & 52 (20.8\%) \\
    \textbf{Thai} & 40 (16.0\%) \\
    \textbf{Chinese} & 43 (17.2\%) \\

    \bottomrule
    \end{tabular}
}
\caption{\textbf{Number of updated MGSM queries per language.} Total count: 250.}
\label{tab:mgsm_details}
\end{table}

%% file: tables/models.tex
\begin{table*}
\centering
\resizebox{\linewidth}{!}{%
    \begin{tabular}{lllp{5cm}p{8cm}}
    \toprule
    \textbf{Model} & \textbf{Context Len.} & \textbf{Vocab. Size} & \textbf{HuggingFace Model Identifier} & \textbf{Train Set} \\
    \toprule

    \textbf{\textsc{Distill-Qwen 1.5b}} & 128K & 152K & \url{deepseek-ai/DeepSeek-R1-Distill-Qwen-1.5B} & Pre-trained on mathematical reasoning queries; post-trained on 70K logical reasoning queries \citep{qwen2025qwen25technicalreport}, further post-trained on reasoning traces distilled from \textsc{DeepSeek-R1} \citep{deepseekai2025deepseekr1incentivizingreasoningcapability}. \\
     
    \textbf{\textsc{Qwen-3 4b}} & 33K & 152K & \url{Qwen/Qwen3-4B} & Pre-trained on reasoning queries; post-trained on a wide range of reasoning tasks \citep{yang2025qwen3technicalreport}. \\

    \bottomrule
    \end{tabular}
}
\caption{\textbf{List of evaluated models.} We report the context length, vocabulary size, and HuggingFace model identifiers. We use \textsc{Qwen-3 4b} with \colorbox{gray!20}{\texttt{enable\_think=True}} mode.}  
\label{tab:model_details}
\end{table*}

%  and \textsc{SmolLM3 3b} with \colorbox{gray!20}{\texttt{\textbackslash think}} mode

%% file: tables/datasets.tex
\begin{table*}
\centering
\resizebox{\linewidth}{!}{%
    \begin{tabular}{llllllllll}
    \toprule
    \textbf{Dataset} & \textbf{\# Queries} & \textbf{Translation} \\
    \toprule

    \textbf{MGSM-Rev2} & 250 & Human-translated by professional translators \citep{shi2022language} \\
    \textbf{MSVAMP} & 250 & Machine translated with \textsc{ChatGPT}\footnote{\url{https://openai.com/index/chatgpt/}}; translation quality verified by native speakers on a random subset \\
    % \textbf{PolyMath} & 500 & Machine translated with \textsc{GPT-4o} \citep{openai2024gpt4ocard}; post-edited by language experts \\

    \bottomrule
    \end{tabular}
}
\caption{\textbf{Detailed statistics of evaluation datasets.}}  
\label{tab:dataset_details}
\end{table*}

% PolyMath has 150 queries for each of four difficulty tiers (low, medium, high, and top), summing up to 500 queries in total.

%% file: tables/languages.tex
\definecolor{midpink}{RGB}{225, 149, 171}
\definecolor{darkpink}{RGB}{186, 67, 101}
\definecolor{darkerpink}{RGB}{99, 36, 54}

\definecolor{midgreen}{RGB}{96, 171, 117}
\definecolor{darkgreen}{RGB}{32, 84, 47}
\definecolor{darkergreen}{RGB}{0, 156, 43}

\definecolor{midblue}{RGB}{66, 135, 245}
\definecolor{darkkblue}{RGB}{35, 77, 145}
\definecolor{darkerblue}{RGB}{96, 117, 150}

\begin{table*}
\centering
\resizebox{\linewidth}{!}{%
    \begin{tabular}{llllllllll}
    \toprule
    \textbf{Language Family} & \textbf{Language} & \textbf{Script} & \textbf{Synthesis} & \textbf{Word Order} & \textbf{Resource Level} & \textbf{\# Speakers} & \textbf{\# Wikipedia Size} \\
    \toprule

    \multirow{6}{*}{Indo-European} & English & Latin & \textcolor{midblue}{analytic} & \textcolor{midpink}{SVO} & \cellcolor{midgreen!40} high & \cellcolor{midgreen!80} 1,130M & \cellcolor{darkgreen!50} 5,758,285 \\
    
    & French & Latin & \textcolor{darkkblue}{fusional} & \textcolor{midpink}{SVO} & \cellcolor{midgreen!40} high & \cellcolor{midgreen!30} 398M & \cellcolor{darkgreen!30} 2,325,608 \\
    
    & Spanish & Latin & \textcolor{darkkblue}{fusional} & \textcolor{midpink}{SVO} & \cellcolor{midgreen!40} high & \cellcolor{midgreen!60} 592M & \cellcolor{darkgreen!20} 1,669,181 \\

    & Portuguese & Latin & \textcolor{darkkblue}{fusional} & \textcolor{midpink}{SVO} & \cellcolor{orange!30} mid & \cellcolor{midgreen!30} 269M & \cellcolor{darkgreen!20} 1,171,437 \\

    & German & Latin & \textcolor{darkkblue}{fusional} & \textcolor{midpink}{SVO}, \textcolor{darkpink}{SOV} & \cellcolor{orange!30} mid & \cellcolor{orange!30} 178M & \cellcolor{darkgreen!30} 2,651,352 \\
    
    & Russian & Cyrillic & \textcolor{darkkblue}{fusional} & \textcolor{midpink}{SVO} & \cellcolor{orange!30} mid & \cellcolor{midgreen!30} 260M & \cellcolor{darkgreen!20} 1,476,045 \\
    
    & Bengali & Bengali & \textcolor{darkkblue}{fusional} & \textcolor{darkpink}{SOV} & \cellcolor{red!20} low & \cellcolor{midgreen!30} 337M & \cellcolor{red!15} 63,762 \\
    \midrule
    
    Sino-Tibetan & Chinese & Chinese & \textcolor{midblue}{analytic} & \textcolor{midpink}{SVO} & \cellcolor{midgreen!40} high & \cellcolor{midgreen!80} 1,350M & \cellcolor{darkgreen!20} 1,246,389 \\
    \midrule
    
    Koreanic & Korean & Hangul & \textcolor{darkerblue}{agglutinative} & \textcolor{darkpink}{SOV} & \cellcolor{orange!30} mid & \cellcolor{red!30} 80M & \cellcolor{orange!15} 437,373 \\
    \midrule

    Japonic & Japanese & Japanese & \textcolor{darkerblue}{agglutinative} & \textcolor{darkpink}{SOV} & \cellcolor{orange!30} mid & \cellcolor{red!20} 128M & \cellcolor{darkgreen!20} 1,133,444 \\
    \midrule
    
    % Afro-Asiatic & Arabic & Arabic & \textcolor{darkkblue}{fusional} & \textcolor{darkerpink}{VSO} & \cellcolor{orange!30} mid & \cellcolor{midgreen!60} 630M & \cellcolor{orange!30} 656,982 \\
    % \midrule
    
    Niger-Congo & Swahili & Latin & \textcolor{darkerblue}{agglutinative} & \textcolor{midpink}{SVO} & \cellcolor{red!20} low & \cellcolor{red!30} 83M & \cellcolor{red!25} 47,793 \\
    \midrule

    Dravidian & Telugu & Telugu & \textcolor{darkerblue}{agglutinative} & \textcolor{darkpink}{SOV} & \cellcolor{red!20} low & \cellcolor{red!30} 96M & \cellcolor{red!15} 66,353 \\
    \midrule

    Kra-Dai & Thai & Thai & \textcolor{midblue}{analytic} & \textcolor{midpink}{SVO} & \cellcolor{red!20} low & \cellcolor{red!30} 72M & \cellcolor{orange!15} 128,179 \\
    
    \bottomrule
    \end{tabular}
}
\caption{\textbf{Characteristics of tested languages.} For each language, we show language family, script, linguistic typologies (synthesis and word order), and resource level measured by the number of speakers and Wikipedia articles \citep{zhang-etal-2023-miracl}.}
\label{tab:lang_details}
\end{table*}

%% file: tables/lang_consistency.tex
\begin{table*}
\centering
\resizebox{\linewidth}{!}{%
 \begin{tabular}{l p{13cm} rr}
 \toprule
 \textbf{Language} & \textbf{Reasoning trace} & \textbf{Language Consistency (↑)} & \textbf{LCPR (\%, ↑)} \\
 \toprule

    \textbf{German} & \colorbox{yellow!30}{Let's solve the problem:} Um denursprünglichen Preis des Buches zu finden, Divisionstypen mit einem Rabatt von 25\% auf den ursprünglichen Preisize. \colorbox{yellow!30}{First,} wirten wir die Gleichung auf, um denursprünglichen Preis (p) zu finden: p - 0.25p = 19.50. \colorbox{yellow!30}{This simplifies to:} 0.75p = 19.50. \colorbox{yellow!30}{Solving for p,} wir p durch Division beider Seiten durch 0.75: p = \textbackslash \textbackslash\{19.50\}\{0.75\}. \colorbox{yellow!30}{Calculating this gives:} p = 26. \textbackslash \textbackslash boxed\{26\} & 1.0 & 0.0 \\
    \midrule 

    \textbf{Korean} & 1. 월요일에 먹은 쿠키 수: 5개 2. 화요일에 먹은 쿠키 수: 13개 3. \colorbox{yellow!30}{기대ly,} 13개 - 5개 = 8개의 쿠키를 \colorbox{yellow!30}{더 did} 4. \colorbox{yellow!30}{하 Alto에 4배 더 information:} 8개 × 4 = 32개 5. \colorbox{yellow!30}{하 Alto에 2배 더 information:} 32개 × 2 = 64개 6. 하 Alto에 1개의 쿠키는 200칼로리: 64개 × 200칼로리 = 12,800칼로리 7.  \colorbox{yellow!30}{summaries:} \textbackslash \textbackslash boxed\{12,800\}칼로리 \colorbox{yellow!30}{difference} & 1.0 & 0.0 \\
    \midrule
 
    \textbf{Japanese} & \begin{CJK*}{UTF8}{min}
   まず、問題を理解します。問題は、青いネクタイの2倍の数の赤いネクタイを\CJKfamily{gbsn}买\CJKfamily{min}います。
   赤いネクタイは青いネクタイよりも50\%高いです。
   彼はそれぞれ\$40の青いネクタイに\$200使いました。
   彼はネクタイに\CJKfamily{gbsn}多少钱\CJKfamily{min}使いましたか？
   まず、青いネクタイの数を\CJKfamily{gbsn}设定\CJKfamily{min}します。
   \colorbox{yellow!30}{Let's denote}青いネクタイの数をCとします。
   \CJKfamily{gbsn}紧接着\CJKfamily{min}、赤いネクタイの数は青いネクタイの2倍です。
   \colorbox{yellow!30}{Therefore,} 赤いネクタイの数は2Cです。
   青いネクタイの価値はPとします。\CJKfamily{gbsn}根据题目\CJKfamily{min}、R = 0.5Pです。
   P = \$40です。R = \$200です。
   \CJKfamily{gbsn}换句话说\CJKfamily{min}、合計はC × \$40 + 2C × \$200 = \$440Cです。
   \colorbox{yellow!30}{\CJKfamily{gbsn}我们可以得出结论。}\CJKfamily{min}
   \textbackslash\textbackslash boxed\{440C\}
   \end{CJK*} & 1.0 & 0.0 \\

 \bottomrule
 \end{tabular}
}
\caption{\textbf{Example of reasoning traces with language consistency and LCPR metric scores.} A reasoning trace can receive a language consistency score of 1.0 despite \colorbox{yellow!30}{substantial code-switching,} which LCPR correctly penalizes.}  
\label{tab:lang_consistency}
\end{table*}

%% file: tables/code-switch.tex
\begin{table}[H]
\centering
\resizebox{\linewidth}{!}{%
    \begin{tabular}{lrrr}
    \toprule
    \textbf{Model} & \textbf{$\text{CSR}_\text{in}$} & \textbf{$\text{CSR}_\text{out}$} &
    \textbf{$\text{CSR}_\text{all}$} \\
    \toprule

    Vanilla & 52.0 & 59.1 & 56.7 \\
    Prompt & 48.7 & 49.2 & 49.0 \\
    \textit{M-Thinker Iter1} & 77.3 & 66.7 & 70.2 \\
    \textit{M-Thinker Iter2} & 92.1 & 76.9 & 82.0 \\
    SFT & 18.9 & 42.3 & 34.6 \\
    +GRPO & 19.1 & 44.0 & 35.4 \\
    +AdaMame-GRPO & 16.8 & 41.9 & 33.2 \\

    \bottomrule
    \end{tabular}
}
\caption{\textbf{Code-switching rate (CSR).} Backbone model: \textsc{Distill-Qwen 1.5b}; Dataset: MGSM-Rev2.}  
\label{tab:code-switch}
\end{table}

%% file: tables/length.tex
\begin{table*}
\centering
\resizebox{\linewidth}{!}{%
    \begin{tabular}{lrrrrrr}
    \toprule
    \textbf{Length bin}
    & \textbf{SFT training data} 
    & \textbf{Vanilla}
    & \textbf{M-Thinker Iter2}
    & \textbf{SFT}
    & \textbf{+GRPO}
    & \textbf{+AdaMame-GRPO} \\
    \toprule

    $<$500
    & 5,250 (17.9\%)
    & 592 (59.2\%)
    & 0 (0.0\%)
    & 878 (87.8\%)
    & 860 (86.0\%)
    & 881 (88.1\%) \\
    
    500--750
    & 7,331 (25.0\%)
    & 233 (23.3\%)
    & 0 (0.0\%)
    & 87 (8.7\%)
    & 94 (9.4\%)
    & 90 (9.0\%) \\
    
    750--1k
    & 6,432 (22.0\%)
    & 82 (8.2\%)
    & 1 (0.1\%)
    & 22 (2.2\%)
    & 29 (2.9\%)
    & 14 (1.4\%) \\
    
    1k--2k
    & 9,547 (32.6\%)
    & 40 (4.0\%)
    & 60 (6.0\%)
    & 7 (0.7\%)
    & 15 (1.5\%)
    & 15 (1.5\%) \\
    
    2k--4k
    & 703 (2.4\%)
    & 13 (1.3\%)
    & 326 (32.6\%)
    & 1 (0.1\%)
    & 1 (0.1\%)
    & 0 (0.0\%) \\
    
    4k+
    & 4 (0.0\%)
    & 40 (4.0\%)
    & 613 (61.3\%)
    & 5 (0.5\%)
    & 1 (0.1\%)
    & 0 (0.0\%) \\
    \midrule
    \textbf{Avg.}
    & 903.4
    & 1,260.3
    & 5,936.6
    & 411.0
    & 317.2
    & 297.4 \\
    
    \bottomrule
    \end{tabular}
}
\caption{\textbf{Length distributions of SFT training data (30K) and baseline traces.} We measure length only for the in-domain languages. Dataset: MGSM-Rev; Model: \textsc{Distill-Qwen 1.5B}.}  
\label{tab:length}
\end{table*}

%% file: tables/change_in_b.tex
\begin{table}[htp]
\centering
\resizebox{\linewidth}{!}{%
    \begin{tabular}{lrrr}
    \toprule
    \textbf{Dataset} & \textbf{$\beta$} & \textbf{Accuracy (\%, ↑)} & \textbf{LCPR (\%, ↑)} \\
    \midrule
    \rowcolor{gray!20}
    \multicolumn{4}{c}{\textbf{\textsc{Distill-Qwen 1.5b}}} \\
    \midrule

    MGSM-Rev2 & 1 & 68.0 & 69.8 \\
     & 2 & 67.9 & 70.1 \\
     & 3 & 66.2 & 70.5 \\
     & 4 & 59.8 & 71.1 \\
     & 5 & 59.3 & 71.6 \\
    \midrule
    MSVAMP & 1 & 77.7 & 59.6 \\
    & 2 & 77.0 & 60.7 \\
    & 3 & 76.6 & 60.8 \\
    & 4 & 76.3 & 61.2 \\
    & 5 & 75.7 & 61.6 \\
    \midrule

    \rowcolor{gray!20}
    \multicolumn{4}{c}{\textbf{\textsc{Qwen3 4b}}} \\
    \midrule
    MGSM-Rev2 & 1 & 86.6 & 87.4 \\
    & 2 & 85.9 & 89.9 \\
    & 3 & 85.4 & 90.1 \\
    & 4 & 80.5 & 90.4 \\
    & 5 & 79.0 & 90.7 \\
    \midrule
    MSVAMP & 1 & 90.1 & 78.7 \\
    & 2 & 89.0 & 86.0 \\
    & 3 & 88.8 & 87.4 \\
    & 4 & 88.4 & 87.9 \\
    & 5 & 86.9 & 88.0 \\

    \bottomrule
    \end{tabular}
}
\caption{\textbf{Numerical results for varying query alignment factor $\beta$.} Increasing \(\beta\) improves LCPR, at the cost of lower reasoning accuracy.}  
\label{tab:change_in_b}
\end{table}

%% file: tables/per_lang_1.tex
\begin{table*}
\centering
\resizebox{\linewidth}{!}{%
    \normalsize
    \begin{tabular}{ll *{18}{r}}
    \toprule
    \textbf{Model} & \textbf{Size} & \textbf{Metric} & \textbf{fr} & \textbf{ja} & \textbf{ko} & \textbf{th} & \textbf{bn} & \textbf{en} & \textbf{es} & \textbf{ru} & \textbf{sw} & \textbf{te} & \textbf{zh} & \textbf{de} & \textbf{Avg.} \\

    \midrule
    \rowcolor{gray!20}
    \multicolumn{16}{c}{\textbf{\textsc{Distill-Qwen 1.5b}}} \\
    \midrule
    Vanilla & - & Acc. & 70.4 & 48.0 & 50.4 & 18.8 & 32.0 & 91.2 & 72.0 & 70.0 & 3.2 & 8.8 & 89.6 & 55.6 & 50.8 \\
    Prompt & - & Acc. & 74.8 & 53.2 & 52.4 & 26.0 & 44.0 & 96.8 & 73.6 & 77.2 & 8.4 & 7.6 & 89.6 & 64.8 & 55.7 \\
    \textit{M-Thinker Iter1} & 35K & Acc. & 84.0 & 67.2 & 48.8 & 48.0 & 47.6 & 96.8 & 87.2 & 82.8 & 9.6 & 17.2 & 93.2 & 79.2 & 63.5 \\
    \textit{M-Thinker Iter2} & 50K & Acc. & 86.8 & 72.4 & 62.8 & 56.0 & 50.8 & 95.2 & 85.6 & 83.6 & 11.2 & 19.6 & 94.0 & 80.0 & 66.5 \\
    SFT & 30K & Acc. & 77.6 & 60.4 & 61.6 & 61.6 & 42.4 & 90.0 & 80.4 & 73.2 & 6.8 & 14.8 & 86.0 & 72.4 & 60.6 \\
    +GRPO & 35K & Acc. & 83.2 & 68.0 & 70.0 & 69.2 & 46.8 & 94.4 & 84.0 & 80.4 & 15.6 & 22.4 & 88.0 & 76.8 & 66.6 \\
    +AdaMame-GRPO & 35K & Acc. & 85.2 & 68.4 & 70.0 & 68.0 & 49.2 & 96.4 & 84.8 & 80.8 & 15.6 & 25.2 & 91.2 & 80.4 & 67.9 \\
    \midrule
    Vanilla & - & LCPR & 92.5 & 33.8 & 17.3 & 64.0 & 16.0 & 90.0 & 94.8 & 24.0 & 28.8 & 47.5 & 92.5 & 88.5 & 57.5 \\
    Prompt & - & LCPR & 91.5 & 37.6 & 37.1 & 65.0 & 14.6 & 68.9 & 92.5 & 34.3 & 18.8 & 60.5 & 90.7 & 88.3 & 58.3 \\
    \textit{M-Thinker Iter1} & 35K & LCPR & 0.0 & 0.0 & 1.0 & 0.2 & 3.8 & 57.4 & 0.8 & 3.2 & 6.0 & 1.2 & 78.0 & 1.8 & 12.8 \\
    \textit{M-Thinker Iter2} & 50K & LCPR & 0.2 & 0.0 & 0.0 & 0.0 & 4.5 & 54.4 & 0.0 & 0.0 & 0.8 & 4.0 & 77.3 & 0.8 & 11.8 \\
    SFT & 30K & LCPR & 90.3 & 70.3 & 74.8 & 75.0 & 49.6 & 81.6 & 87.4 & 16.9 & 49.0 & 47.8 & 93.2 & 77.0 & 67.7 \\
    +GRPO & 35K & LCPR & 88.6 & 68.4 & 78.6 & 84.8 & 29.2 & 81.2 & 85.2 & 18.0 & 54.0 & 52.0 & 93.4 & 79.7 & 67.8 \\
    +AdaMame-GRPO & 35K & LCPR & 89.3 & 69.1 & 78.4 & 85.5 & 52.4 & 81.5 & 87.8 & 18.3 & 57.4 & 50.2 & 92.8 & 78.0 & 70.1 \\
    \midrule
    Vanilla & - & TTC & 6.8 & 19.0 & 12.2 & 60.7 & 26.3 & 9.4 & 4.7 & 14.2 & 37.5 & 38.3 & 10.4 & 6.3 & 20.5 \\
    Prompt & - & TTC & 5.2 & 28.7 & 11.3 & 72.5 & 53.2 & 24.0 & 7.6 & 42.9 & 59.7 & 64.5 & 17.2 & 8.6 & 33.0 \\
    \textit{M-Thinker Iter1} & 35K & TTC & 39.5 & 19.2 & 17.2 & 24.0 & 15.9 & 24.8 & 33.6 & 36.1 & 16.4 & 11.4 & 16.0 & 32.5 & 23.9 \\
    \textit{M-Thinker Iter2} & 50K & TTC & 26.7 & 19.3 & 17.3 & 20.4 & 14.5 & 18.4 & 32.5 & 23.2 & 19.9 & 13.9 & 13.3 & 33.0 & 21.0 \\
    SFT & 30K & TTC & 1.6 & 0.8 & 1.5 & 2.1 & 3.8 & 1.6 & 2.8 & 3.0 & 13.4 & 6.3 & 0.7 & 2.4 & 3.3 \\
    +GRPO & 35K & TTC & 1.3 & 0.1 & 1.0 & 1.7 & 2.6 & 1.2 & 1.5 & 1.5 & 5.9 & 3.6 & 0.7 & 1.8 & 1.9 \\
    +AdaMame-GRPO & 35K & TTC & 1.3 & 1.0 & 0.9 & 1.6 & 2.3 & 1.3 & 1.5 & 1.7 & 5.6 & 3.1 & 0.7 & 1.9 & 1.9 \\

    \midrule
    \rowcolor{gray!20}
    \multicolumn{16}{c}{\textbf{\textsc{Qwen3 4b}}} \\
    \midrule

    Vanilla & - & Acc. & 98.0 & 70.0 & 64.8 & 92.4 & 88.4 & 99.6 & 99.2 & 94.8 & 37.6 & 78.0 & 32.8 & 88.4 & 78.7 \\
    Prompt & - & Acc. & 98.4 & 86.8 & 83.2 & 83.6 & 84.8 & 98.8 & 99.6 & 94.4 & 30.0 & 77.6 & 68.4 & 89.2 & 82.9 \\
    SFT & 30K & Acc. & 94.0 & 91.2 & 84.4 & 89.6 & 83.6 & 96.0 & 96.4 & 93.6 & 36.8 & 73.2 & 93.6 & 95.2 & 85.6 \\
    +GRPO & 35K & Acc. & 94.8 & 90.0 & 82.0 & 88.8 & 82.4 & 98.0 & 94.4 & 94.0 & 34.8 & 71.6 & 93.2 & 94.4 & 84.9 \\
    +AdaMame-GRPO & 35K & Acc. & 96.0 & 90.0 & 84.8 & 90.8 & 81.6 & 96.4 & 95.6 & 94.4 & 38.8 & 73.2 & 94.4 & 94.4 & 85.9 \\
    \midrule
    Vanilla & - & LCPR & 0.6 & 1.6 & 1.6 & 1.4 & 2.2 & 76.2 & 0.8 & 90.7 & 7.7 & 4.9 & 93.2 & 0.2 & 23.4 \\
    Prompt & - & LCPR & 0.8 & 1.2 & 1.8 & 1.8 & 4.1 & 79.8 & 1.0 & 91.7 & 13.8 & 4.3 & 92.9 & 0.2 & 24.5 \\
    SFT & 30K & LCPR & 93.9 & 78.1 & 89.0 & 90.4 & 91.8 & 87.3 & 93.0 & 92.6 & 79.3 & 93.3 & 94.7 & 79.8 & 88.6 \\
    +GRPO & 35K & LCPR & 94.9 & 76.4 & 89.7 & 92.2 & 91.4 & 86.8 & 93.5 & 92.7 & 79.8 & 92.5 & 94.8 & 90.5 & 89.6 \\
    +AdaMame-GRPO & 35K & LCPR & 96.1 & 78.3 & 89.3 & 91.8 & 91.8 & 86.5 & 93.6 & 92.8 & 80.2 & 93.7 & 94.3 & 90.7 & 89.9 \\
    \midrule
    Vanilla & - & TTC & 13.4 & 14.7 & 13.9 & 15.1 & 17.8 & 22.4 & 13.6 & 13.2 & 34.3 & 19.2 & 7.8 & 13.9 & 16.6 \\
    Prompt & - & TTC & 12.5 & 14.4 & 13.2 & 15.9 & 18.4 & 20.0 & 13.3 & 12.2 & 38.4 & 19.2 & 8.4 & 10.9 & 16.4 \\
    SFT & 30K & TTC & 1.1 & 0.7 & 0.8 & 7.6 & 1.2 & 1.2 & 1.2 & 1.2 & 2.7 & 2.1 & 0.6 & 1.2 & 1.8 \\
    +GRPO & 35K & TTC & 1.1 & 0.7 & 0.8 & 5.6 & 1.3 & 1.2 & 1.3 & 1.2 & 2.6 & 1.6 & 0.6 & 1.3 & 1.6 \\
    +AdaMame-GRPO & 35K & TTC & 1.1 & 0.7 & 0.8 & 4.9 & 1.2 & 1.2 & 1.1 & 1.1 & 2.6 & 1.7 & 0.6 & 1.3 & 1.5 \\

    \bottomrule
    \end{tabular}
}

\caption{\textbf{Per-language results for MGSM-Rev2 dataset.}}
\label{tab:per_lang_1}
\end{table*}

%% file: tables/per_lang_2.tex
\begin{table*}
\centering
\resizebox{\linewidth}{!}{%
    \normalsize
    \begin{tabular}{ll *{18}{r}}
    \toprule
    \textbf{Model} & \textbf{Size} & \textbf{Metric} & \textbf{fr} & \textbf{ja} & \textbf{th} & \textbf{bn} & \textbf{en} & \textbf{es} & \textbf{sw} & \textbf{ru} & \textbf{zh} & \textbf{de} & \textbf{Avg.} \\

    \midrule
    \rowcolor{gray!20}
    \multicolumn{14}{c}{\textbf{\textsc{Distill-Qwen 1.5b}}} \\
    \midrule
    Vanilla & - & Acc. & 82.4 & 71.6 & 45.6 & 48.0 & 93.2 & 88.0 & 14.8 & 85.2 & 90.0 & 84.8 & 70.4 \\
    Prompt & - & Acc. & 84.4 & 73.6 & 42.8 & 56.8 & 92.0 & 88.4 & 16.8 & 86.8 & 89.6 & 83.2 & 71.4 \\
    \textit{M-Thinker Iter1} & 35K & Acc. & 86.2 & 74.4 & 76.0 & 56.0 & 94.8 & 86.4 & 21.0 & 88.8 & 82.0 & 80.0 & 74.6 \\
    \textit{M-Thinker Iter2} & 50K & Acc. & 86.4 & 78.8 & 78.0 & 57.4 & 96.4 & 90.3 & 20.4 & 90.8 & 86.8 & 80.4 & 76.6 \\
    SFT & 30K & Acc. & 83.2 & 76.4 & 72.8 & 57.6 & 94.8 & 91.6 & 26.8 & 82.6 & 88.8 & 85.6 & 76.0 \\
    +GRPO & 35K & Acc. & 84.8 & 74.8 & 72.4 & 55.6 & 94.8 & 92.4 & 31.6 & 83.6 & 88.0 & 87.6 & 76.6 \\
    +AdaMame-GRPO & 35K & Acc. & 84.0 & 76.0 & 72.4 & 58.8 & 94.4 & 90.8 & 29.6 & 87.8 & 88.8 & 87.6 & 77.0 \\
    \midrule
    Vanilla & - & LCPR & 95.0 & 35.8 & 61.5 & 16.5 & 99.1 & 94.3 & 20.9 & 28.2 & 90.7 & 1.6 & 54.4 \\
    Prompt & - & LCPR & 96.2 & 36.1 & 66.3 & 9.90 & 98.8 & 94.1 & 39.4 & 28.1 & 92.8 & 1.6 & 56.3 \\
    \textit{M-Thinker Iter1} & 35K & LCPR & 8.1 & 0.6 & 0.6 & 19.2 & 81.7 & 2.7 & 0.0 & 14.9 & 87.7 & 1.0 & 21.7 \\
    \textit{M-Thinker Iter2} & 50K & LCPR & 3.1 & 1.0 & 0.4 & 11.5 & 81.6 & 4.3 & 1.0 & 6.3 & 86.4 & 1.6 & 19.7 \\
    SFT & 30K & LCPR & 89.4 & 76.3 & 86.4 & 52.1 & 93.3 & 84.3 & 8.80 & 17.9 & 92.6 & 1.4 & 60.3 \\
    +GRPO & 35K & LCPR & 91.2 & 78.7 & 84.6 & 29.3 & 94.0 & 81.7 & 8.60 & 23.9 & 93.3 & 1.2 & 58.7 \\
    +AdaMame-GRPO & 35K & LCPR & 91.8 & 77.4 & 86.3 & 54.2 & 94.3 & 83.2 & 9.00 & 16.6 & 92.5 & 1.2 & 60.7 \\
    \midrule
    Vanilla & - & TTC & 3.6 & 8.4 & 17.0 & 7.7 & 3.1 & 3.3 & 13.5 & 6.3 & 6.3 & 3.6 & 7.3 \\
    Prompt & - & TTC & 3.8 & 8.7 & 19.6 & 8.0 & 3.0 & 3.3 & 16.8 & 5.6 & 2.3 & 3.8 & 7.5 \\
    \textit{M-Thinker Iter1} & 35K & TTC & 24.3 & 11.6 & 16.1 & 13.7 & 10.2 & 19.6 & 12.8 & 20.0 & 6.9 & 16.4 & 15.2 \\
    \textit{M-Thinker Iter2} & 50K & TTC & 18.9 & 11.1 & 14.1 & 10.8 & 9.8 & 19.4 & 16.8 & 15.5 & 6.7 & 18.8 & 15.2 \\
    SFT & 30K & TTC & 1.2 & 0.7 & 1.1 & 1.9 & 0.9 & 0.8 & 4.8 & 1.2 & 1.0 & 1.3 & 1.5 \\
    +GRPO & 35K & TTC & 0.8 & 0.8 & 1.3 & 2.0 & 0.8 & 0.9 & 4.4 & 1.3 & 0.4 & 1.1 & 1.4 \\
    +AdaMame-GRPO & 35K & TTC & 0.8 & 0.7 & 1.4 & 1.6 & 0.9 & 1.2 & 4.6 & 0.9 & 0.5 & 1.1 & 1.4 \\

    \midrule
    \rowcolor{gray!20}
    \multicolumn{14}{c}{\textbf{\textsc{Qwen3 4b}}} \\
    \midrule

    Vanilla & - & Acc. & 93.2 & 46.4 & 88.4 & 53.2 & 86.4 & 83.6 & 75.6 & 53.2 & 18.0 & 71.6 & 67.0 \\
    Prompt & - & Acc. & 91.8 & 70.2 & 89.0 & 60.5 & 86.4 & 86.4 & 81.2 & 62.4 & 92.4 & 84.0 & 80.4 \\
    SFT & 30K & Acc. & 89.6 & 92.0 & 85.6 & 80.4 & 94.4 & 95.2 & 90.4 & 66.8 & 93.2 & 92.0 & 88.0 \\
    +GRPO & 35K & Acc. & 90.0 & 92.8 & 87.6 & 81.6 & 94.0 & 94.8 & 90.4 & 69.2 & 92.8 & 92.8 & 88.6 \\
    +AdaMame-GRPO & 35K & Acc. & 92.4 & 91.6 & 88.0 & 81.6 & 94.0 & 95.2 & 91.6 & 70.4 & 92.8 & 92.4 & 89.0 \\
    \midrule
    Vanilla & - & LCPR & 0.0 & 0.2 & 0.6 & 0.6 & 91.9 & 0.2 & 91.0 & 0.6 & 93.3 & 0.8 & 27.9 \\
    Prompt & - & LCPR & 0.0 & 0.0 & 1.0 & 0.6 & 91.1 & 0.4 & 91.5 & 2.0 & 93.5 & 0.8 & 28.1 \\
    SFT & 30K & LCPR & 94.2 & 88.8 & 92.6 & 97.3 & 97.0 & 92.0 & 95.4 & 2.3 & 91.8 & 91.6 & 84.3 \\
    +GRPO & 35K & LCPR & 96.2 & 89.4 & 92.3 & 96.7 & 97.2 & 93.0 & 96.0 & 2.0 & 92.9 & 94.2 & 85.0 \\
    +AdaMame-GRPO & 35K & LCPR & 96.5 & 90.2 & 93.3 & 96.9 & 97.9 & 94.7 & 96.6 & 3.9 & 94.8 & 94.7 & 86.0 \\
    \midrule
    Vanilla & - & TTC & 10.7 & 10.5 & 12.7 & 14.4 & 16.8 & 10.0 & 12.0 & 24.6 & 7.5 & 11.7 & 13.1 \\
    Prompt & - & TTC & 10.9 & 10.4 & 13.6 & 16.9 & 17.8 & 10.5 & 11.6 & 26.7 & 8.0 & 9.7 & 13.6 \\
    SFT & 30K & TTC & 1.1 & 0.7 & 1.6 & 0.9 & 1.0 & 1.4 & 0.9 & 2.6 & 0.5 & 1.2 & 1.2 \\
    +GRPO & 35K & TTC & 0.8 & 0.5 & 1.1 & 1.0 & 0.8 & 0.9 & 1.0 & 2.2 & 0.4 & 1.0 & 1.0 \\
    +AdaMame-GRPO & 35K & TTC & 0.8 & 0.5 & 0.9 & 0.9 & 0.8 & 0.9 & 0.9 & 1.8 & 0.3 & 0.9 & 0.9 \\

    \bottomrule
    \end{tabular}
}

\caption{\textbf{Per-language results for MSVAMP dataset.}}
\label{tab:per_lang_2}
\end{table*}